\documentclass[12pt]{article}
\usepackage{amsmath}
\usepackage{graphicx}
\usepackage{enumerate}
\usepackage{natbib}
\usepackage{xr-hyper}
\usepackage{url} % not crucial - just used below for the URL
\usepackage{hyperref}
\hypersetup{breaklinks=true}
\usepackage{xcolor}
\usepackage{bm,bbm}
\usepackage{amsbsy,amsfonts,amssymb}
\usepackage{enumitem}
\usepackage{pdfpages}
\usepackage{algorithm}
\usepackage{algpseudocode}
\usepackage{subcaption}
\usepackage{soul, mathtools}
\usepackage{titlesec}
\usepackage{environ}
\usepackage{array}
\usepackage{multirow}
\usepackage{authblk}
\setcitestyle{aysep={}}

\newcommand{\blind}{0}

\titleformat{\subsubsection}[runin]% runin puts it in the same paragraph
       {\normalfont\bfseries}% formatting commands to apply to the whole heading
       {\thesubsubsection}% the label and number
       {0.5em}% space between label/number and subsection title
       {}% formatting commands applied just to subsection title
       []% punctuation or other commands following subsection title

\newcommand{\mystrut}{\vphantom{\int_0^1}}

\newtheorem{Def}{Definition}
\newtheorem{Lem}{\underline{\bf Lemma}}
\newtheorem{Pro}{Proposition}
\newtheorem{Th}{\underline{\bf Theorem}}

\newtheorem{Rem}{\underline{\bf Remark}}

\newcommand{\defeq}{\vcentcolon=}

\def\trans{^{\top}}
\def\wh{\widehat}
\def\wt{\widetilde}

\def\argmin{\mbox{argmin}}

\def\0{{\bf 0}}
\def\1{{\bf 1}}

\def\a{{\bf a}}
\def\b{{\bf b}}
\def\c{{\bf c}}
\def\d{{\bf d}}
\def\e{{\bf e}}
\def\p{{\bf p}}
\def\t{{\bf t}}
\def\u{{\bf u}}
\def\v{{\bf v}}
\def\x{{\bf x}}
\def\z{{\bf z}}

\def\A{{\bf A}}
\def\B{{\bf B}}
\def\C{{\bf C}}
\def\D{{\bf D}}

\def\M{{\bf M}}
\def\S{{\bf S}}
\def\W{{\bf W}}
\def\X{{\bf X}}
\def\Z{{\bf Z}}

\def\bR{\mathbb{R}}
\def\mC{\mathcal{C}}

\def\mF{\mathcal{F}}
\def\mG{\mathcal{G}}
\def\mH{\mathcal{H}}
\def\mL{\mathcal{L}}
\def\mN{\mathcal{N}}
\def\mS{\mathcal{S}}

\def\ba{\boldsymbol\alpha}
\def\bb{{\boldsymbol\beta}}
\def\bg{{\boldsymbol\gamma}}

\def\beps{{\boldsymbol\epsilon}}
\def\bpsi{{\boldsymbol\psi}}

\def\sumi{\sum_{i=1}^n}

\def\be{\begin{eqnarray}}
\def\ee{\end{eqnarray}}
\def\bse{\begin{eqnarray*}}
\def\ese{\end{eqnarray*}}

\def\boxit#1{\vbox{\hrule\hbox{\vrule\kern6pt
\vbox{\kern6pt#1\kern6pt}\kern6pt\vrule}\hrule}}

\begin{document}

\def\spacingset#1{\renewcommand{\baselinestretch}%
{#1}\small\normalsize} \spacingset{1}

%%%%%%%%%%%%%%%%%%%%%%%%%%%%%%%%%%%%%%%%%%%%%%%%%%%%%%%%%%%%%%%%%%%%%%%%%%%%%%

\if0\blind
{
  \title{\bf Nested Deep Learning Model\\Towards a Foundation Model for Brain Signal Data}
  \author[1]{Fangyi Wei}
  \author[2]{Jiajie Mo}
  \author[2]{Kai Zhang}
  \author[1]{Haipeng Shen}
  \author[3]{Srikantan Nagarajan}
  \author[4]{Fei Jiang\thanks{Email: fei.jiang@ucsf.edu}}
  \affil[1]{Innovation and Information Management, Faculty of Business and Economics, The University of Hong Kong, China}
  \affil[2]{Department of Neurosurgery, Beijing Tiantan Hospital, China}
  \affil[3]{Department of Radiology, University of California, San Francisco}
  \affil[4]{Department of Epidemiology and Biostatistics, University of California, San Francisco}
%   \author{Fangyi Wei \\
%     Innovation and Information Management,
%     Faculty of Business and Economics, \\
%     The University of Hong Kong, China \\
%         Jiajie Mo \\
%     Department of Neurosurgery,
%     Beijing Tiantan Hospital, China \\
%          Kai Zhang \\
%     Department of Neurosurgery,
%     Beijing Tiantan Hospital, China \\
%  Haipeng Shen \\
%     Innovation and Information Management,
%     Faculty of Business and Economics, \\
%     The University of Hong Kong, China \\
% Srikantan Nagarajan \\
%     Department of Radiology,
%     University of California, San Francisco  \\
%     Fei Jiang\thanks{Email: fei.jiang@ucsf.edu} \\
%     Department of Epidemiology and Biostatistics, \\
%     University of California, San Francisco  \\
%     }
    \date{}
%  \author{Author 1\thanks{
%    The authors gratefully acknowledge \textit{please remember to list all relevant funding sources in the unblinded version}}\hspace{.2cm}\\
%    Department of YYY, University of XXX\\
%    and \\
%    Author 2 \\
%    Department of ZZZ, University of WWW}
  \maketitle
} \fi

\if1\blind
{
  \bigskip
  \bigskip
  \bigskip
  \title{\bf Nested Deep Learning Model\\Towards a Foundation Model
    for Brain Signal Data}
    \author{}
    \date{}
  
  \maketitle
  
  \medskip
  \bigskip
} \fi

% \bigskip
\begin{abstract}
Epilepsy affects around 50 million people globally. Electroencephalography (EEG) or Magnetoencephalography (MEG) based spike detection plays a crucial role in diagnosis and treatment. Manual spike identification is time-consuming and requires specialized training that further limits the number of qualified professionals. To ease the difficulty, various algorithmic approaches have been developed. However, the existing methods face challenges in handling varying channel configurations and in identifying the specific channels where the spikes originate. A novel Nested Deep Learning (NDL) framework is proposed to overcome these limitations. NDL applies a weighted combination of signals across all channels, ensuring adaptability to different channel setups, and allows clinicians to identify key channels more accurately. Through theoretical analysis and empirical validation on real EEG/MEG datasets, NDL is shown to improve prediction accuracy, achieve channel localization, support cross-modality data integration, and adapt to various neurophysiological applications.

\end{abstract}

\noindent%
\vfill

%%%%%%%%%%%%%%%%%%%%%%%%%%%%%%%%%%%%%%%%%%%%%%%%%%%%%%%%%%%%%%%%%%%%%%%%%%%%%%

\newpage
\spacingset{1.9} % DON'T change the spacing!
\section{Introduction}
\label{sec:intro}

Currently, around 50 million people worldwide suffer from epilepsy, one of the most common neurological diseases globally \citep{WHOepilepsy}.
Detecting epileptiform discharges, particularly spikes, during the interictal periods, plays a crucial role in the diagnosis of epilepsy.

Spikes are abnormal brain activities that typically occur intermittently and spontaneously, lasting around 100 to 250 milliseconds, in epilepsy patients.
They can be measured with Electroencephalography (EEG) or Magnetoencephalography (MEG) devices.
The devices collect brain signals through EEG or MEG sensors attached to different areas of the scalp and visualize the signals as time series waveforms.
The EEG or MEG recordings then consist of multi-channel time series signal data.
Identifying spikes not only helps predict the risk of seizures, but also helps localize the onset zone of seizures, which is critical for epilepsy surgery \citep{smith2022human}.

Thus, it becomes imperative to develop novel interpretable methods that can effectively identify spikes to assist clinical decisions. However, clinical practices predominantly employ manual identification of spikes, which is a labor-intensive process, especially when analyzing continuous or video EEG studies that can last more than 24 hours.
Furthermore, this task requires specialized training, restricting the number of clinicians qualified to evaluate EEG studies, resulting in a significant shortage of neurologists trained in interpreting EEG data \citep{torres1996impact}.

To mitigate this issue, various algorithm-based methods have been proposed to automate the identification process.
\citet{indiradevi2008multi} use wavelet transformation to decompose spike signals.
\citet{lodder2013inter} adopt template matching to compare signal patterns with spike templates.
\citet{baud2018unsupervised} use an unsupervised learning approach to automate spike detection.
%\citet{fukumori2021epileptic} introduce a linear-phase filter for spike detection.
With the advancement of deep learning, the field has increasingly embraced deep learning algorithms for spike detection.
For example,
\citet{antoniades2016deep}, \citet{johansen2016epileptiform}, \citet{tjepkema2018deep}, \citet{thomas2020automated}, \citet{clarke2021computer}, \citet{chung2023deep} and \citet{munia2023interictal} employ convolutional neural networks (CNNs) to classify abnormal and normal waveforms;
\citet{medvedev2019long} use a Long Short-Term Memory neural network for the spike detection task.

Despite these developments, two significant challenges remain.
Firstly,  different studies have varying numbers and locations of the channels, as well as the montages used, all of which make it difficult to develop a universal algorithm that can accommodate all channel setups.
For instance, \citet{golmohammadi2019automatic} develop an algorithm based on spikes labeled using the temporal central parasagittal montage with 22 channels, whereas \citet{jing2020development}'s algorithm is designed for the longitudinal bipolar montage with 18 channels and the common average reference montage with 19 channels. 
A general adaptive algorithm is crucial because it allows data integration between studies, thereby enhancing prediction accuracy.
It also allows clinicians to choose their preferred montage for analysis.

Secondly, although current detection algorithms can identify time segments that contain spikes, they lack the ability to pinpoint the exact channels where the spikes originate.
Identifying specific channels is crucial because it helps locate the seizure onset zone, helps clinicians review and validate detection results, and serves as a key factor in ruling out false detections in less likely channels.

We propose a novel nested deep learning framework (NDL) to address these challenges.
For the first challenge, we propose a weight function for each channel, customized to its specific signal pattern within a given time range of interest, preferably less than one second.
Then we model the probability of a time segment to contain a spike via a weighted combination of signals across all channels.
For the second challenge, these weights are designed to reflect the importance of each channel in terms of its relevance in the spike detection.

Our NDL approach is superior to the traditional channel-based methods \citep{wilson1999spike, goelz2000wavelet}, which require the tedious process of labeling each channel to achieve similar results.
NDL offers a faster inference process that can handle large volumes of data simultaneously.
Furthermore, NDL considers the signal information from all channels, which improves the prediction accuracy because there are often concurrent spikes that appear on multiple channels.
Moreover, because NDL is invariant to channel configurations, it can serve as a foundational model for integrating neurophysiological data from various modalities and can be fine-tuned to adapt to different new tasks.

We apply NDL to three real datasets to evaluate its performance in Section~\ref{sec:real}.
We start with a well-labeled EEG dataset from Temple University Hospital (TUH), focusing on its sensitivity and specificity in detecting spikes, as well as its accuracy in identifying the EEG channels where these spikes occur (Section~\ref{subsec:tuh}).
The dataset is the only public EEG dataset that provides not only the time label of spike occurrence but also the specific channels containing the spikes, as the ground truth for the evaluation.
We then apply NDL to two private datasets: a 140-channel MEG dataset from the University of California, San Francisco (UCSF), and a 19-channel EEG dataset from China's Beijing Tiantan Hospital (BTH).
The results first demonstrate that the model trained on the UCSF MEG data performs well in detecting spikes and identifying relevant channels (Section~\ref{subsec:ucsf}).
Furthermore, the MEG-trained model can even successfully detect spikes in the out-of-sample BTH EEG data, even though the testing data modality differs from the training data modality (Section~\ref{subsec:bth}).
This study indicates that NDL has strong generalizability across different datasets, even with varying numbers of channels.
Finally, Section~\ref{subsec:continuous} proposes an algorithm that enables the application of NDL to continuous MEG/EEG recordings commonly available in practice.

The rest of the paper is presented as follows.
We introduce the three real datasets and their preprocessing procedures in Section~\ref{sec:data}.
Then we introduce the model in Section~\ref{sec:meth}, and present the deep learning estimation procedure and the asymptotic consistency in Section~\ref{sec:est}.
Finally, we demonstrate the performance of NDL on the three real datasets and its comparison with existing methods in Section~\ref{sec:real}.
% we demonstrate its performance and convergence on simulated data in Section~\ref{sec:simu}.
% Finally, we study the theoretical properties of the estimator in Section~\ref{sec:th}.

To facilitate the presentation, we define the following notations.
% The sets of natural numbers, natural numbers including 0 and real numbers are denoted by $\mathbb{N}, \mathbb{N}_0$ and $\mathbb{R}$, respectively.
The $L_0$, $L_1$, $L_2$ and the supremum norms of a vector $\v$ are denoted by $\|\v\|_0$, $\|\v\|_1$, $\|\v\|_2$ and $\|\v\|_{\infty}$, respectively.
Furthermore, for a given matrix $\M$, we denote the matrix entry-wise supremum norm, $L_0$, $L_1$,  the operator norms as $\|\M\|_{\infty}$, $\|\M\|_0$, $\|\M\|_1$, and $\|\M\|_2$, respectively.
For a function $f(\x)$, we denote the functional supremum norm as
$\|f(\x)\|_{\infty} \equiv \sup |f(\x)|$.

\vspace{3mm}

{Note:
The replication codes are provided: \url{https://github.com/shirleyweify/NDL_Replication_Code.git}.
The open data model is made publicly available, whereas access to the private data model requires a formal request.}

\vspace{-3mm}
%%%%%%%%%%%%%%%%%%%%%%%%%%%%%%%%%%%%%%%%%%%%%%%%%%%%%%%%%%%%%%%%%%%%%%%%%%%%%%
\section{Three Real Datasets}
\label{sec:data}

\vspace{-5mm}
\subsection{TUH Dataset}
\label{subsec:tuhdata}

The TUH dataset \citep{harati2015improved} is the only publicly available scalp interictal EEG dataset that includes per-channel annotations.
The dataset consists of 511 continuous EEG records using the standard 10/20 channel configuration, collected from 370 patients.
Each record is collected at a sampling frequency of 250 Hz.
We filter the data and retain frequencies between 1 and 70 Hz to remove the slow wave and fast oscillations caused by the artifacts and noise.
Then we remove the linear trend of the data.
The data are then transformed into the Temporal Central Parasagittal montage, as described in \citet{lopez2016analysis}, using the 22 channels listed in Table~\ref{tab:TCP} in the supplementary material.
This montage was chosen because it had been used when labeling the data.

Each EEG record has one or more channel-specific one-second annotations.
According to \citet{obeid2016temple}, these annotations include six classes:
(1) spike and sharp wave (SPSW),
(2) generalized periodic epileptiform discharge (GPED),
(3) periodic lateralized epileptiform discharge (PLED),
(4) eye movement (EYEM),
(5) artifact (ARTF), and
(6) background (BCKG).
Among the six classes, 
SPSW, GPED and PLED are abnormal events of clinical interest, while EYEM, ARTF and BCKG are artifacts and normal signals.

\vspace{2mm}
\begin{figure}[htbp]
\begin{center}
\includegraphics[width=1.0\textwidth]{fig/viz\_eval.pdf}
\end{center}
\vspace{-5mm}
\caption{TUH: Two-second segments of the six classes of events with 22 ACNS TCP montages, where the one-second annotated events are shown in red and bold. \label{fig:TUHwaveform}}
\vspace{-2mm}
\end{figure}

We extract a two-second segment around each annotation, where the middle one-second segment is annotated, while the remaining is treated as background references.
In total, we obtain 12,675 EEG segments from the TUH dataset.
% is treated as $\X_i$ and the remaining forms $\Z_i$.
% Both $\X_i$ and $\Z_i$ are $22\times 250$.
Figure~\ref{fig:TUHwaveform} provides examples of the two-second segments for each class of the annotations.
Table~\ref{tab:channelinfo} reports the number of segments and the number of patients for each class.

\begin{table}[htbp]
\caption{TUH: Numbers of segments and corresponding numbers of patients for each of the six classes, in both the training set and the testing set.
\label{tab:channelinfo}}
\vspace{-3mm}
\begin{center}
  \begin{tabular}[width=1.0\textwidth]{cc|ccc|ccc}
    \hline
    \multicolumn{2}{c|}{TUH Data}
   & \multicolumn{3}{|c|}{Abnormal}&\multicolumn{3}{c}{Normal}\\
\hline
\multicolumn{2}{c|}{}
& {SPSW} & {GPED} & {PLED}
& {EYEM} & {ARTF} & {BCKG} \\
\hline
  {Training} & \multicolumn{1}{|c|}{\# Segments} & 144 & 2607 & 2072 & 498 & 1587 & 3232 \\
  & \multicolumn{1}{|c|}{\# Patients} & 35 & 50 & 49 & 73 & 165 & 264 \\
\hline
  {Testing} & \multicolumn{1}{|c|}{\# Segments} & 41 & 616 & 516 & 146 & 371 & 845 \\
  & \multicolumn{1}{|c|}{\# Patients} & 23 & 49 & 46 & 49 & 123 & 236 \\
\hline
\end{tabular}
\end{center}
\vspace{-10mm}
\end{table}

\subsection{UCSF MEG Dataset}
\label{subsec:ucsfmegdata}

The USCF MEG data are recorded from 277 epilepsy patients who have undergone MEG tests at the MEG center of UCSF between 2019 and 2022.
Each MEG recording has a labeling file stored on the clinical server.
First, the recording is fully screened by a technologist to mark out candidate spikes.
These candidate spikes are then reviewed by a clinician, often with 1 or 2 residents, to confirm the presence of the spikes.
This process ensures that the spike labels are validated through multiple readers.

The MEG records are collected at a sampling frequency of 600 Hz on 140 channels.
We filter the data to retain frequencies between 1 and 45 Hz, downsample the data to 256 Hz, and remove the linear trend of the data.
Since the spike labels are less than 250 milliseconds, we extract a segment of 0.5 seconds around each label, where the middle 0.25-second segment contains a spike annotation and the remaining signals are background references.
% Both $\X_i$ and $\Z_i$ are $140\times 64$.
We also extract 0.5-second background segments from non-spike-labeled time.
These background segments can accurately represent non-spike activities.
We obtain 230,325 segments, among which 46\% are spike segments labeled in the clinical notes.

\subsection{BTH EEG Dataset}
\label{subsec:btheegdata}

The BTH EEG data are collected at China's Beijing Tiantan Hospital, to test the adaptability of NDL across multi-modal data.
% The dataset consists of 106 EEG segments with 19 channels from 20 patients, with 53 segments containing spike signals.
The dataset consists of 84 EEG records with 19 channels, collected from 20 patients.
In total, 53 spike events are annotated and confirmed by two trained clinicians.
To increase the variability in background waveforms, we extract 50 subsets of background events from the normal periods.
Each sub-dataset then consists of the 53 spike segments and another 53 randomly-selected background segments.

The raw data are collected at a sampling frequency of 512 Hz.
We apply a filter to retain frequencies between 1 and 45 Hz, downsample the data to 256 Hz, and remove the linear trend of the data.
The data are then converted into the common average montage.
% The spikes are labeled and confirmed by two trained clinicians. 
We extract 0.5-second time segments around the labeled spikes and normal background activities, where the middle 0.25-second segment is of our interest and the remaining serves as reference signals.
% Both $\X_i$ and $\Z_i$ are $19\times 64$.
Finally, we apply the pre-trained model from the UCSF MEG data on the out-of-sample EEG data, to evaluate NDL's adaptability across different datasets.

%%%%%%%%%%%%%%%%%%%%%%%%%%%%%%%%%%%%%%%%%%%%%%%%%%%%%%%%%%%%%%%%%%%%%%%%%%%%%%
\section{Model}
\label{sec:meth}

\begin{figure}[!htbp]
\begin{center}
\includegraphics[width=1.0\textwidth]{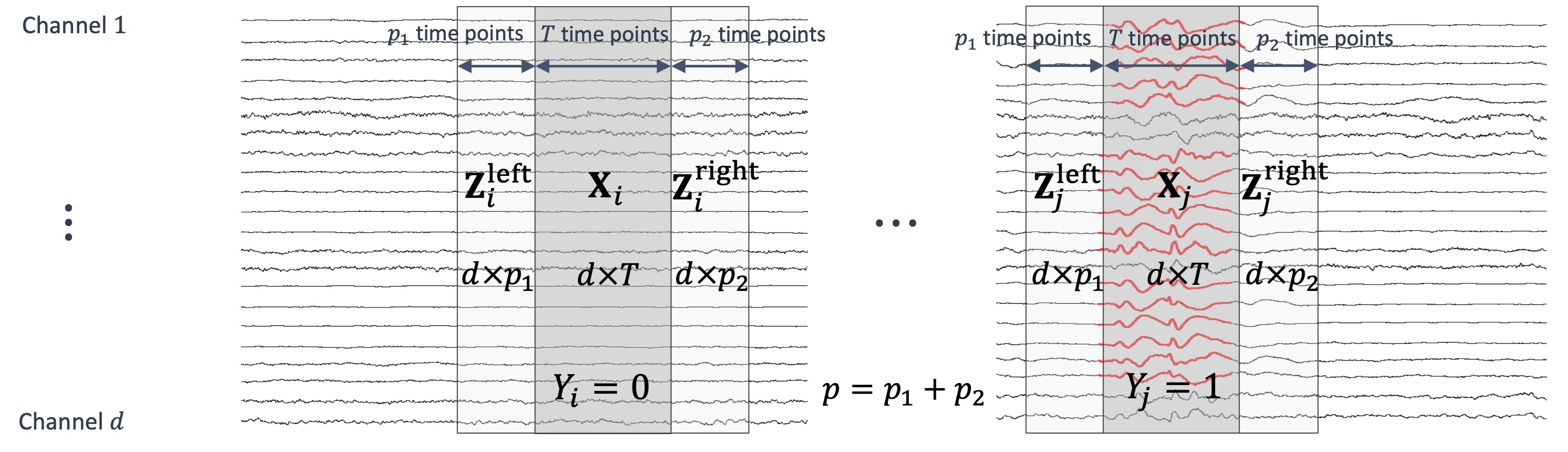}
\end{center}
\vspace{-5mm}
\caption{Illustration of $\{Y_i=0, \X_i, \Z_i\}$ (left) and $\{Y_j=1, \X_j, \Z_j\}$ (right).
Spike signals are annotated in red and bold while background signals are shown in black.
In the TUH data of Section~\ref{subsec:tuhdata}, we set $d=22$, $T=p=250$, $p_1=p_2=125$.
\label{fig:sample}}
\vspace{-2mm}
\end{figure}

Let $\X_i = \{\X_{il}, l = 1, \ldots, d\}^\top$ be a random matrix representing a segment of multi-channel signals, where $d$ is the number of channels, $\X_{il}$ is a $T$-dimensional vector representing the signals in Channel $l$ with $T$ time measurements.
We aim to predict whether the segment $\X_i$ contains a spike or not.
Let $Y_i$ be a binary indicator, where $Y_i = 1$ indicates that a spike appears in at least one channel in the segment $\X_i$, and $Y_i=0$ indicates that no spike appears in the segment.
Furthermore, let $\Z_i = \{\Z_{il}, l = 1, \ldots, d\}^\top$ represent a segment of background signals surrounding $\X_i$, where $\Z_{il}$ is a $p$-dimensional vector containing signals from $p$ time measurements in Channel $l$.
We assume that $\{Y_i, \X_i, \Z_i\},\; i = 1, \ldots, n$, are independent identically distributed.
Figure~\ref{fig:sample} provides two exemplary segments where $Y_i = 0$ and $Y_j = 1$, with the corresponding $\{\X_i, \Z_i\}$ and $\{\X_j, \Z_j\}$.

We define a channel-specific weight function as $\ba(\X_{il}, \X_i) = \{\alpha_k(\X_{il}, \X_i), k = 1, \ldots, p\}\trans$, where $\alpha_k(\X_{il}, \X_i) = \exp\{\omega_k(\X_{il})\} / \sum_{l=1}^d \exp\{\omega_k(\X_{il})\}$ represents the $k$-th weight, dynamically standardized across $d$ channels.
Here, $\omega_k(\X_{il}),\; k = 1, \ldots, p$, are unknown real-valued functions.
The weight function $\ba(\X_{il}, \X_i)$ incorporates information from all channels while placing particular emphasis on Channel $l$, highlighting the importance of signal patterns in that channel.
Then we combine the signals including $\X_i$ and $\Z_i$, through  
\be\label{eq:Sa}
\S_i(\ba) \equiv \sum_{l=1}^d \left\{
\X_{il} \ba(\X_{il}, \X_i)\trans + \1_T \ba(\X_{il}, \X_i)\trans \Z_{il} \1_p^\top
\right\},
\ee
where $\1_T$ and $\1_p$ are $T$- and $p$- dimensional one vectors, respectively.
We introduce an unknown real function $g (\cdot): \bR^{T\times p} \to \bR$ to capture the complex relationship between the combination defined in (\ref{eq:Sa}) and the response $Y_i$.
We then assume that the conditional probability for $Y_i = 1$ is in the following form: 
\be\label{eq:logmodel}
\Pr\left(Y_i =1| \X_i, \Z_i\right) 
&=& \left\{ 1+ \exp[-g \{\S_i(\ba)\}] \right\}^{-1}, 
\ee
where $g$ and $\ba$ are unknown functions that need to be estimated.

%A formal proof of the proposition is presented in Section~\ref{sec:pro:pro1} in the supplementary material.
\begin{Pro}\label{pro:1}
Assume $g(\cdot)$ is continuously differentiable, $\ba(\X_{il}, \X_i)>0$ entry-wise and $\sum_{l=1}^{d} \ba(\X_{il}, \X_i) = \bf 1$.
Then $\ba(\X_{il}, \X_i)$ and  $g(\cdot)$ are identifiable.
\end{Pro}
\begin{Rem}
% We make some remarks about the formulation (\ref{eq:Sa}) and Model (\ref{eq:logmodel}).
Assuming $\sum_{l=1}^{d} \ba(\X_{il}, \X_i) = \bf 1$ is essential for the identifiability of the model. Based on this assumption, we propose the softmax-standardized form of $\ba$.
In the formulation (\ref{eq:Sa}),
the term $\sum_{l=1}^d \X_{il} \ba(\X_{il}, \X_i) \trans$ maps the signals $\X_i$ from  $d$ channels onto a shared $p$-dimensional space. 
Intuitively, a higher value of $\|\ba(\X_{il}, \X_i)\|_2$ indicates a greater contribution of Channel $l$ to the outcome.
It is worth noting that using only this term makes the model unidentifiable because the order of the columns of $\ba(\X_{il}, \X_i)$ can vary arbitrarily.
The term $\sum_{l=1}^d \1_T \ba(\X_{il}, \X_i)\trans \Z_{il} \1_p^\top$ is introduced to address this problem.
By multiplying each element of $\ba(\X_{il}, \X_i)$ with a distinct element of $\Z_{il}$, we effectively fix the order of the columns of $\ba(\X_{il}, \X_i)$, ensuring the identifiability of the model.
% We show that
\end{Rem}

In the model,  the dimension $p$, i.e. the dimension of $\Z_{il}$ and the dimension of $\ba(\X_{il}, \X_i)$, should be selected appropriately depending on the context.
When $p<T$, a low-dimensional embedding of the data is obtained, which can be beneficial for dimension reduction in analyses involving a large number of time points.
This approach is particularly useful for tasks such as seizure detection or resting-state signal classification, where signal recordings span across minutes or hours.
In contrast,  when $p = T$, all the original signal information is preserved.
This choice is recommended for tasks like spike detection, where the analysis involves only short data segments, typically containing less than 100 time points.
We set $p=T$ for the real data analyses in this paper.
% Note that $p>T$ is not used as it introduces redundant information and imposes a heavier computational burden.
We select symmetric intervals $\Z_i^{\text{left}}$ and $\Z_i^{\text{right}}$ primarily for convenience. Moreover, by definition, a spike must stand out against its surrounding background, so we sample background segments around each identified spike. Since there is no clinical evidence that pre‑spike and post‑spike backgrounds differ in informational content, we deliberately choose a symmetric background window to avoid introducing any bias.

Our proposal is related to, but distinct from, both the attention mechanism \citep{vaswani2017attention} and the Mixture-of-Experts model \citep{jacobs1991adaptive}.
We define a weight function $\ba(\X_{il}, \X_i)$ to represent channel importance as a function of the signal itself.
While this resembles the key-query weighting in attention and the weighted combination in Mixture-of-Experts, our method introduces structural constraints on the weight function, which preserve model identifiability and enhance interpretability.

\section{Deep Learning Estimation}\label{sec:est}

Deep neural networks (DNNs) have emerged as powerful tools for estimating unknown functions in the nonparametric regression.
\citet{yarotsky2017error} and \citet{imaizumi2019deep} demonstrate that deep neural networks (DNNs) outperform other models in approximating unknown nonparametric functions.
Furthermore, \citet{schmidt2020nonparametric} provides comprehensive analyses showing that deep networks achieve minimax optimal rates for estimating H\"older smooth functions.
To flexibly capture the structures of the unknown functions $g$ and $\ba$ in Model (\ref{eq:logmodel}), we adopt DNNs with the ReLU activation function \citep{schmidt2020nonparametric} to approximate  the underlying true functions $g^*$ and $\ba^*$.
% More specifically, let $g^*$ and $\ba^*$ denote the true functions for $g$ and $\ba$ in Model (\ref{eq:logmodel}), respectively.
% To flexibly capture the structures of the unknown functions, we adopt DNNs with the ReLU activation function \citep{schmidt2020nonparametric}
% to approximate $g^*$ and $\ba^*$.

More specifically, let network architectures with $L$ numbers of layers and a width vector $\p_{w}$, where $\p_{w} = (p_{w0}, \ldots, p_{w (L+1)}) \in \mathbb{N}^{L + 1}$, represent any function in the form of
\be\label{eq:NN}
f: \mathbb{R}^{p_{w0}} \to \mathbb{R}^{p_{w (L + 1)}}, 
\x \to f(\x) \equiv 
\W_L \sigma_{\v_L} \W_{L-1} \sigma_{\v_{L-1}}, \ldots, \W_1 \sigma_{\v_1}\W_0 \x, 
\ee
where $\sigma_{\v}(y_1, \ldots, y_r) = \left\{\sigma(y_1- v_1), \ldots, \sigma(y_r-v_r)\right\}$ is a vector of ReLU activation functions, $\W_i$ is a $p_{w (i+1)} \times p_{wi}$ weight matrix, and $\v_i \in \bR^{p_{wi}}$ is a shift vector.
Then we denote the set of these neural network functions with $s$ numbers of nonzero parameters as
\bse
\mF(L, \p_w, s, F)
&\equiv&
\left\{ 
f \text{ of the form (\ref{eq:NN})}:
\max_{j = 0, \ldots, L} \max(\|\W_j\|_{\infty}, \|\v_j\|_{\infty}) \leq 1,
\right. \\
&&
\left.
\sum_{j=1}^L \|\W_j\|_0 + \|\v_j\|_0\leq s,
\sup_{\x} |f(\x)| \leq F
\right\}. 
\ese
Denote $\mF_{f} = \mF(L_f, \p_f, s_f, F_f)$, where $f$ can be $g, \omega_k, k = 1, \ldots, p$.

Furthermore, we define a general form of the true function space as follows.  \begin{Def}\label{def:PC}
Define the set  of $\beta$-H\"older functions with radius $K$ as
\bse
\mC_t^\beta(D, K)
& \equiv &
\left\{
f: D \subset \bR^t \to \bR:
\right. \\
&& \left.
\sum_{\bg: \|\bg\|_1 <\beta} \|\partial^{\bg} f\|_{\infty} 
+ \sum_{\bg: \|\bg\|_1 = \lfloor\beta \rfloor}
\sup_{\x_1, \x_2 \in D, \x_1\neq \x_2}
\frac{|\partial^{\bg} f(\x_1) - \partial^{\bg} f(\x_2)|}
{\|\x_1 - \x_2\|_{\infty}^{\beta - \lfloor\beta \rfloor}} 
\leq K
\right\},
\ese
where $\partial^{\bg} = \partial^{\gamma_1}\ldots \partial^{\gamma_t}$ with $\bg = (\gamma_1, \ldots, \gamma_t)\trans \in \mathbb{N}^t$, and $\lfloor\beta \rfloor$ denotes the largest integer strictly smaller than $\beta$.
Then the set of true function space is defined as 
\bse
\mG(q, \d, \t, \bb, K)
&\equiv&
\left\{
f = g_q\circ \ldots \circ g_0:
g_u \equiv (g_{uj})_j: [a_u, b_u]^{d_u} \to [a_{u+1}, b_{u + 1}]^{d_{u + 1}},
\right.\\
&& \left.
g_{uj} \in \mC_{t_u}^{\beta_u}([a_u, b_u]^{t_u}, K),
\text{ for some } |a_i|, |b_i| <K
\right\}, 
\ese
where $\d \equiv (d_0, \ldots, d_{q+1})$, $\t\equiv (t_0, \ldots, t_q)\trans$ and $\bb \equiv (\beta_0, \ldots, \beta_q)\trans$.
\end{Def} 
For notational convenience, we denote the true function space for an arbitrary function $f$ as $\mG_f \equiv \mG(q_f, \d_f, \t_f, \bb_f, K_f)$.
Assume that $g^* \in \mG_g, \omega^*_k \in \mG_{\omega_k}$ are the true functions of $g, \omega_k$.
Let $\alpha_k^*(\X_{il}, \X_i) = \exp\{\omega_k^*(\X_{il})\} / \sum_{l=1}^d \exp\{\omega_k^*(\X_{il})\}$ and $\ba^*(\X_{il}, \X_i) = \{\alpha_k^*(\X_{il}, \X_i), k = 1, \ldots, p\}^\top$.
We can estimate $g^*, \ba^*$ by minimizing the negative log-likelihood
\be\label{eq:loss}
&\mL(g, \ba)=-n^{-1}
\sumi\left(
Y_i g\left\{\S_i(\ba)\right\} - h\left[g\left\{\S_i(\ba)\right\}\right]
\right)
\text{ subject to } \nonumber\\
& g \in \mF_g,  \omega_k \in \mF_{\omega_k},
\alpha_k(\X_{il}, {\X_i}) = \exp\{\omega_k(\X_{il})\} / \sum_{l=1}^d\exp\{\omega_k(\X_{il})\}, \\
& \ba(\X_{il},{\X_i}) =
\{\alpha_k(\X_{il}, {\X_i}), k = 1, \ldots, p\}\trans \nonumber
\ee
where $h(x) = \log\{1 + \exp(x)\}$.
Let $\wh{g}, \wh{\ba}$ be the resulting estimators.

In practice,  we employ the network architecture depicted in Figure~\ref{fig:dnn} to approximate the true functions $g^*, \ba^*$. As shown in Figure~\ref{fig:dnn}, the core structure consists of a series of recursive CNN layers followed by a fully connected layer.
We use Blocks 1 and 2 to construct functions in $\mF_{w_k},\; k = 1, \ldots, p$, which approximate $\ba^*$. The resulting function is then fed into Blocks 3 and 4 to build a function in $\mF_{g}$ that
approximates $g^*$.

\begin{figure}[!htbp]
\begin{center}
\includegraphics[width=1.0\textwidth]{fig/NDL\_DNN\_highres.drawio.png}
\end{center}
\vspace{-5mm}
\caption{
The DNN diagram to approximate the true functions $g^*, \ba^*$. \label{fig:dnn}}
\vspace{-2mm}
\end{figure}

We establish the asymptotic consistency of the estimators under the Conditions \ref{con:c1}--\ref{con:RE}, which are presented in Section~\ref{sec:supp:cond} in the supplementary material. 
Here,  we first introduce some necessary definitions.
Let $g_0 \in \mF_g, \omega_{0k} \in \mF_{\omega_k}, k = 1, \ldots, p$ such that
$g_0 \equiv \argmin_{g \in \mF_g} \|g - g^*\|_\infty$, and $\omega_{0k} \equiv \argmin_{\omega_k \in \mF_{\omega_k}} \|\omega_k - \omega^*_k\|_\infty$,
where $g^*, \omega_k^*$ are the true functions in $\mG_g, \mG_{\omega_k}$, respectively.
Let $\alpha_{0k}(\X_{il}, {\X_i}) =\exp\{\omega_{0k}(\X_{il})\}/\sum_{l=1}^d\exp\{\omega_{0k}(\X_{il})\}$ and $\ba_0 = \{\alpha_{0k}, k = 1, \ldots, p\}\trans$.
Let $\psi_g = g - g_0$ and $\psi_{\alpha_k} = \alpha_k - \alpha_{0k}$ for arbitrary functions $g\in \mF_g$ and $\alpha_k \in \mH_{\alpha_k}$, where
\bse
\mH_{k} \equiv \left\{\alpha_k =
  \exp\{\omega_k(\X_{il})\}/\sum_{l=1}^d\exp\{\omega_k(\X_{il})\}:
  \omega_k \in \mF_{\omega_k}, k = 1, \ldots, p\right\}.
\ese

\begin{Lem}\label{lem:hh}
% Assuming Conditions \ref{con:c1}--\ref{con:h}
Assuming Conditions \ref{con:c1}--\ref{con:c5} in the supplementary material hold, and $T, p \leq s_g$, there are positive constants $C_g, C_{\omega_j}, c_g, c_{\omega_j}$ such that, with probability one,
\bse
&&
|h'\left[g^*\left \{\S_i(\ba^*) \right\}\right] -
h'\left[g_0\left \{\S_i(\ba_0) \right\}\right]|
\leq
\left\{
C_g \max_{u = 0, \ldots, q_g}
c_g^{-\frac{\beta_{gu}^*}{t_{gu}}}
n^{-\frac{ \beta_{gu}^*}{2\beta_{gu}^* + t_{gu}}}
\right. \\
&& \left. 
+ 2C_MM_L \sqrt{d \min(s_g, Tp)}
\sup_{j=1, \ldots, p} C_{\omega_j}
\max_{u = 0, \ldots, q_{\omega_j}}
c_{\omega_j}^{-\frac{\beta_{\omega_j u}^*}{t_{\omega_j u}}}
n^{-\frac{\beta_{\omega_j u}^*}{2\beta_{\omega_j u}^* + t_{\omega_j u}}}
\right\}. 
\ese
\end{Lem}
The proof of Lemma~\ref{lem:hh} is presented in Section~\ref{sec:pro:lemm1} in the supplementary material.
Note that $h'(\cdot)$ is the expit function: $h'\left[g^*\left \{\S_i(\ba^*) \right\}\right]$ represents the true probability of $Y_i=1$.
Lemma~\ref{lem:hh}  shows that the distance between the deep learning approximation of the success probability and the truth depends on how well $g_0$ and $\ba_0$ approximate the true values.
This distance decreases at the standard nonparametric rate as the sample size grows.

\begin{Th}\label{th:1}
Let $\psi_{\wh g}= \wh{g} - g_0$ and $\psi_{\wh \alpha_j} = \wh{\alpha}_j - \alpha_{0j}$.
Assume $\wh{\alpha}_j, \alpha_{0j} \in \mH_{j}$, $\omega^*_j \in \mG_{\omega_j}$,  $\wh{g}, g_0 \in \mF_g, $ and $g^* \in \mG_{g}$.
Assume $T, p<s_g$ and Conditions \ref{con:c1}--\ref{con:RE} in the supplementary material hold.
Let $d{P_{\ba} (\cdot)}$ be the probability density function of $\S_i(\ba)$, and assume
$\|d{P_{\ba_0} (\z)}/d{P_{\ba^*} (\z)}\|_{\infty}\geq m_h>0$.
Furthermore, let $dP_{\X}(\cdot)$ be the probability density for $\X_{i}$.
When $n\to \infty$ and $T, p$ are finite, we have
\bse
&&
\int \{\wh{g} (\z) - g^* (\z)\}^2 dP_{\ba^*}(\z)
+ \sum_{j=1}^p \int \sum_{l=1}^d 
\{ \wh{\alpha}_{j}(\x_{l}, {\x}) - \alpha^*_j (\x_{l}, {\x}) \} ^2
dP_{{\X}}({\x})
\overset{p}{\rightarrow}  0.
\ese
\end{Th}
A more comprehensive version of the theorem, including its convergence rate, is presented as Theorem~\ref{th:supp1} in Section~\ref{sec:supp:conv} of the supplementary material.
The proof of the main theorem is provided in Section~\ref{sec:pro:th1}.
To establish the result, we first show in Lemma~\ref{lem:firstorder} that the first derivative of the objective function is upper bounded by a term that vanishes as the sample size increases.
Next, Lemma~\ref{lem:secondorder} demonstrates that the second derivative converges to a positive definite matrix.
Building on these, we prove estimation consistency by combining the first- and second-order rates in Lemma~\ref{lem:hh}.
Additionally, Section~\ref{sec:supp:ex} in the supplementary material provides examples that illustrate special cases of compositional constraints on $\omega_k$ for $k = 1, \ldots, p$, and on $g$.
Section~\ref{sec:simu} provides the convergence on the simulated data.

%%%%%%%%%%%%%%%%%

\section{Real Data Analysis}\label{sec:real}

We apply NDL on the three real datasets described in Section~\ref{sec:data}, including the TUH dataset, the UCSF MEG dataset and the BTH EEG dataset.
We also compare NDL with three existing methods: SpikeDet, which uses template matching \citep{spikedet2016}; a deep learning multi-head CNN model trained on the TUH dataset \citep{munia2023interictal}; and SpikeNet, a deep learning approach developed using a private dataset of 9,571 scalp EEG recordings from Massachusetts General Hospital \citep{jing2020development}.

%%%%%%%%%%%%%%%%%%%%%%%%%%%%%%%%%%%%

%%%
\subsection{Results from the TUH Data}\label{subsec:tuh}

We apply NDL to classify the EEG segments from the TUH data into normal and abnormal waveforms.
In the testing set, the $i$th segment is classified as containing an abnormal waveform if the predicted probability of 
$Y_i=1$ is greater than 0.5. We report the following evaluation metrics,  where  $TP$, $TN$, $FP$, and $FN$ denote true positives, true negatives, false positives and false negatives, respectively.

\begin{enumerate}
%\item {Accuracy:} $= \frac{TP + TN}{TP + TN + FP + TN}$
\item $\text{Sensitivity} \defeq TP / (TP + FN)$
%\item {FPR} $\equiv \frac{FP}{FP + TN}$
\item $\text{Precision} \defeq TP / (TP + FP)$
\item $\text{Specificity} \defeq TN / (TN + FP)$
%\item {FDR:} $=\frac{FP}{FP + TP} = 1 - Precision$
\item $\text{F1 score} \defeq TP / (TP + (FP + FN) / 2)$
\item {PRAUC:} The area under the precision-recall curve
\item {AUC:} The area under the receiver operating characteristic
  (ROC) curve.
\end{enumerate}

The TUH dataset is divided into 80\% for training and 20\% for testing.
We use the mini-batch gradient descent to train the loss function.
The learning rate is stepwise and decays by 0.5 every 10 epochs.
The training loss decreases rapidly in the first 10 epochs and becomes stable after the 40th epoch.

\begin{figure}[htbp]

\begin{subfigure}{.5\textwidth}
\centering
\includegraphics[page=2,width=0.9\linewidth]{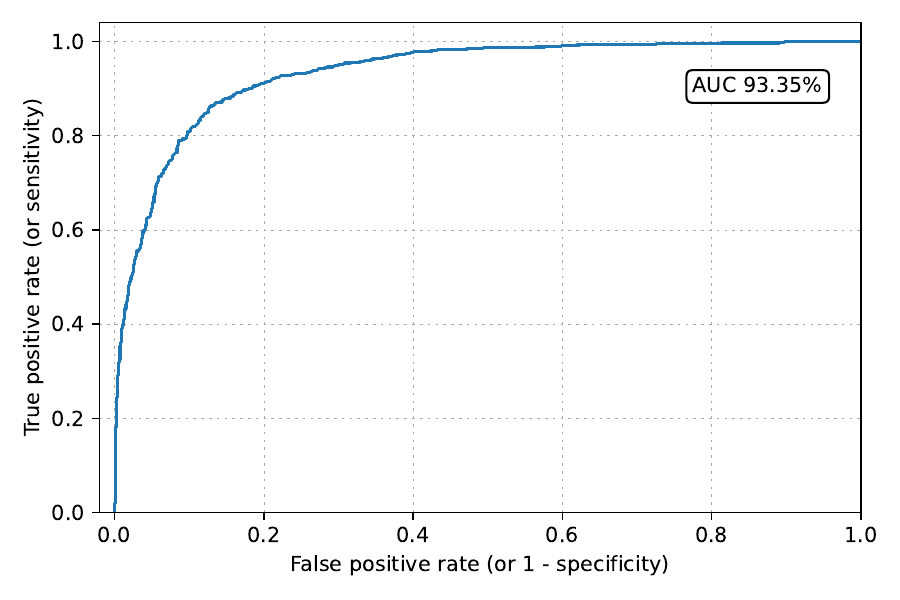}
\vspace{-2mm}
\caption{Test ROC curve}
\end{subfigure}
\begin{subfigure}{.5\textwidth}
\centering
\includegraphics[page=3,width=0.9\linewidth]{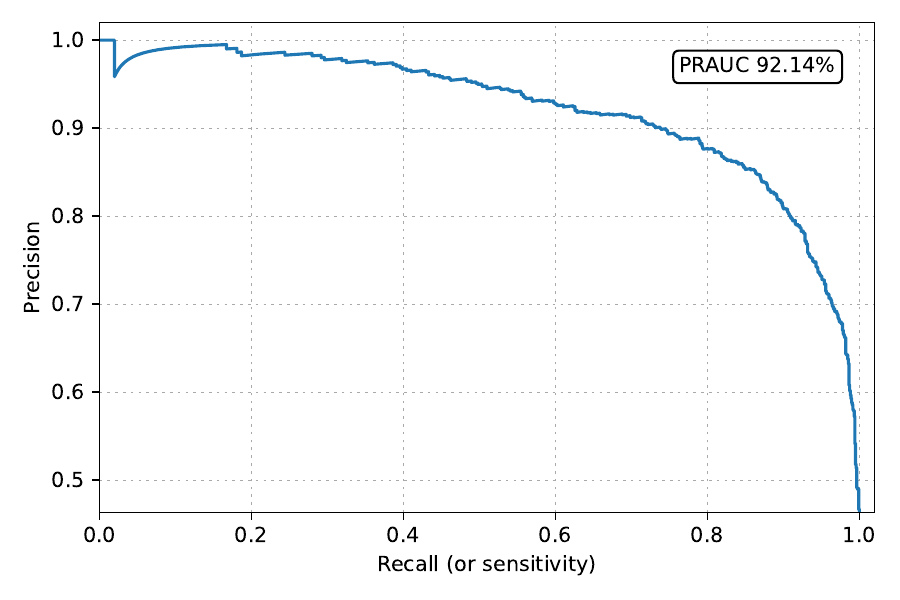}
\vspace{-2mm}
\caption{Test precision-recall curve}
\end{subfigure}

\vspace{-3mm}
\caption{TUH: Test results from NDL with fine-tuned hyperparameters.
\label{fig:TUHres}}
\vspace{-5mm}

\end{figure}

Figures \ref{fig:TUHres} (a) and (b) show the ROC curve and precision-recall curve on the test set. The AUC and PRAUC are 93.35\% and 92.14\%, respectively, demonstrating the high accuracy and precision of the proposed method.
Table~\ref{tab:TUHtest} compares NDL with three competing methods: SpikeDet \citep{spikedet2016}, an open-source spike detection algorithm developed by \citet{nonclercq2009spike}; a deep multi-head CNN model proposed in \citet{munia2023interictal}; and SpikeNet developed by \citet{jing2020development}.  It is worth noting that SpikeNet is a large-scale deep learning model, comprising 318,849 parameters and trained on EEG data from over 19 million EEG segments. As such, it is regarded as the current state-of-the-art benchmark method.

\begin{table}[htbp]
\caption{
% TUH: Comparison of NDL, SpikeNet, the multi-head CNN method in \citet{munia2023interictal}, and SpikeDet. Specificity, PRAUC and AUC are not reported in the multi-head CNN method. AUC and PRAUC are unavailable in SpikeDet.
{TUH: Comparison of NDL, SpikeNet, Multi-head CNN, and SpikeDet.
Specificity, PRAUC and AUC are not reported for Multi-head CNN. PRAUC and AUC are unavailable in SpikeDet, which generates binary outputs rather than probabilities.}
\label{tab:TUHtest}}
\vspace{-5mm}
\begin{center}
\begin{tabular}{ccccccc}
\hline
TUH Data & {Sen (\%)} & {Prec (\%)} & {Spec (\%)} & {F1 (\%)} & {PRAUC (\%)} & {AUC (\%)} \\
\hline
\textbf{NDL} & \textbf{86.96} & {84.51} & {86.27} & \textbf{85.71} & \textbf{92.14} & \textbf{93.35} \\
SpikeNet & {81.33} & \textbf{86.96} & 89.50 & 84.05 & 90.02 & 89.63 \\
Multi-head CNN & {74.71} & {72.70} & - & {73.69} & - & - \\
SpikeDet & {13.30} & {78.39} & \textbf{96.84} & 22.74 & - & - \\
\hline
\end{tabular}
\end{center}
\vspace{-5mm}
\end{table}

Table~\ref{tab:TUHtest} demonstrates that NDL achieves the best performance among all methods, with the highest sensitivity, F1 score, PRAUC, and AUC (highlighted in bold).
In this comparison, NDL, SpikeNet and Multi-head CNN method are deep learning-based methods.
NDL and Multi-head CNN % \citep{munia2023interictal}
are trained and evaluated on the TUH dataset, whereas SpikeNet is pre-trained on one of the largest private EEG datasets \citep{jing2020development} and then tested on the TUH test set.
In contrast, SpikeDet is an unsupervised template matching method, which does not consider spike patterns from noisy data. Therefore, SpikeDet gives the highest number of false negatives among the four methods.

In addition to binary classification, NDL can identify the channel from which the spike likely originates.
Specifically, for each channel $l$,  NDL estimates a p-dimensional weight vector   $\wh \ba(\X_{il}, \X_i)$.
% In addition to binary classification, NDL can estimate the $p$-dimensional weight $\wh \ba(\X_{il}, \X_i)$ for each channel. 
By summing its entries, we obtain an importance score for Channel 
$l$ as $\1\trans \wh \ba(\X_{il}, \X_i)$.
Ranking these scores across all channels allows us to identify the top $L$ channels most likely to contain spikes.

\begin{figure}[htbp]
\begin{center}
\includegraphics[page=1,width=0.95\textwidth]{fig/top\_channels\_tuev.pdf}
\includegraphics[page=2,width=0.95\textwidth]{fig/top\_channels\_tuev.pdf}
\includegraphics[page=3,width=0.95\textwidth]{fig/top\_channels\_tuev.pdf}
\end{center}
\vspace{-5mm}
\caption{
TUH: Side-by-side boxplots comparing the top channels selected by NDL ranking with random ranking through 100 repetitions.
\label{fig:topChannels}}
\vspace{-2mm}
\end{figure}

To assess this approach, we cross-check the selected top channels with the actual channel labels from the TUH data to check whether our model can identify spike-containing channels better than random selection.
Figure~\ref{fig:topChannels} illustrates the probability that the top $L$ channels of a segment contain a true spike, comparing NDL ranking with random ranking based on the TUH testing data. We create the 95\% confidence interval of the probability through resampling procedures with different initial values generated from different random seeds.
It is obvious that NDL performs much better in identifying those channels with spikes. %features that differ from the background noise.
Furthermore, NDL ignores background channels (last pair of side-by-side boxplots). % because the background waveforms represent baseline time series without any distinguishable and clinically significant patterns.

\subsection{Results from the UCSF MEG Data}\label{subsec:ucsf}

We use 95\% of the data for training and the remaining 5\% for testing. Table~\ref{tab:statTEST} summarizes NDL's excellent performance on the testing data, for example, an AUC of 91.61\%.

\begin{table}[htbp]
\caption{
UCSF MEG: Testing Results.
\label{tab:statTEST}}
\vspace{-5mm}
\begin{center}
\begin{tabular}[width=1.0\textwidth]{ccccccc}
\hline
UCSF MEG Data& Sen (\%) & Prec (\%) & Spec (\%) & F1 (\%) & PRAUC (\%) & AUC (\%) \\
\hline
NDL
& 77.07 & 84.26 & 91.39 & 80.50 & 87.20 & 91.61 \\
\hline
\end{tabular}
\end{center}
\vspace{-8mm}
\end{table}

We visualize the distribution of the most important channels using topographical graphs (Figure~\ref{fig:topomap-meg}) and barplots (Figure~\ref{fig:lobeDistribution}), where the topographical graph is a 2D representation of the probabilities of being the top $L$ channels identified by NDL ranking.

\begin{figure}[htbp]
\begin{center}
\includegraphics[page=1,width=0.75\textwidth]{fig/topomap\_meg.pdf}
\end{center}
\vspace{-3mm}
\caption{UCSF MEG: The topological distributions of the probabilities of being the top channels identified by NDL ranking.
\label{fig:topomap-meg}}
\vspace{-5mm}
\end{figure}

Figure~\ref{fig:topomap-meg} contains six topographical graphs, illustrating the probabilities of being identified as one of the most important $L$ channels out of the 140 MEG channels based on the training set and the testing set, respectively.
Figure~\ref{fig:lobeDistribution} shows the distribution of the top $L$ channels across different brain lobes.
Figures \ref{fig:topomap-meg} and~\ref{fig:lobeDistribution} both indicate that the spikes often occur in the temporal lobes. 
%This pattern is observed in both the left and right hemispheres.
This is consistent with the clinical finding that seizures usually happen in the temporal lobes~\citep{mcintosh2019temporal}.

\begin{figure}[htbp]
\begin{center}
\includegraphics[page=3,width=0.95\textwidth]{fig/topomap\_meg.pdf}
\end{center}
\vspace{-3mm}
\caption{UCSF MEG: Histograms of the distributions of the top channels identified by NDL ranking, on five different lobes.
\label{fig:lobeDistribution}}
\vspace{-5mm}
\end{figure}

\subsection{Results from the BTH EEG Data}\label{subsec:bth}

We take the model trained on the UCSF MEG dataset and test it on the BTH EEG dataset. This analysis aims to evaluate the adaptability of NDL across different modalities with varying channels.
% Even without retraining, NDL achieves an AUC of 84.18\% and a sensitivity of 94.34\%.

We evaluate the performance of NDL on 50 randomly selected subsets of the BTH EEG dataset (Section~\ref{subsec:btheegdata}) and compare it against SpikeNet, the strongest competing method.
% Figure~\ref{fig:AUCcomp} presents the distribution of AUC (a) and PRAUC (b) scores for both NDL and SpikeNet.
NDL achieves an AUC ranging from 79.60\% to 90.00\%, a PRAUC from 77.04\% to 91.41\%. In comparison, SpikeNet achieves an AUC between 84.23\% and 93.52\%, a PRAUC between 79.39\% and 95.24\%. Despite being trained on only 230,325 MEG segments and using just 17,683 parameters, NDL achieves performance comparable to the much larger SpikeNet model, which was trained on over 19 million EEG segments and contains more than 300,000 parameters.
Furthermore, these results demonstrate that NDL generalizes well to out-of-sample, cross-modality settings.

% We also evaluate SpikeNet \citep{jing2020development} on this dataset, where it achieves an AUC of 87.75\% and a PRAUC of 82.25\%.
A key advantage of NDL over SpikeNet is its flexibility in handling channel variabilities--NDL remains effective even when certain channels fail to capture brain signals or when additional channels are introduced to improve spatial resolution.
In contrast, SpikeNet becomes unusable when the channel configuration deviates from that of the training data, as it was trained under fixed channel settings.

We present three topographical graphs in Figure~\ref{fig:topomap-tiantan} showing the probability of each channel being selected as one of the top $L$ channels by NDL.
Furthermore, Figure~\ref{fig:lobeDistribution-tiantan} shows the distribution of the top $L$ channels in different lobes.
Interestingly, in addition to the temporal lobes, the prefrontal regions are often selected.
Our neurology collaborators confirm that this is not unexpected, as scalp EEG spikes can often be confused with eye movements, which typically occur in the prefrontal areas.
More data are needed to fine-tune the model for better adaptation to the EEG data.

\begin{figure}[htbp]
\begin{center}
\includegraphics[page=2,width=0.8\textwidth]{fig/topomap\_tiantan.pdf}
\end{center}
\vspace{-5mm}
\caption{BTH EEG: The topological distributions of the probabilities of being selected into the top channels NDL ranking.
\label{fig:topomap-tiantan}}
\vspace{0mm}
\end{figure}

\begin{figure}[htbp]
\begin{center}
\includegraphics[page=3,width=0.9\textwidth]{fig/topomap\_tiantan.pdf}
\end{center}
\vspace{-5mm}
\caption{BTH EEG: Histograms of the distributions of the top channels selected by NDL ranking, on six different lobes.
\label{fig:lobeDistribution-tiantan}}
\vspace{-3mm}
\end{figure}

\subsection{Procedures to adopt to continuous time series data}\label{subsec:continuous}

Unlike the segmented data analyzed in the previous subsections, neurophysiology data are usually recorded continuously in real-world settings.
We develop the following Algorithm~\ref{al:1} that applies NDL to continuous time series data in practice.

Algorithm~\ref{al:1} employs a sliding window technique to extract overlapping data segments and evaluate the probability of containing a spike for each segment.
We then identify those time segments where the probability exceeds a specified threshold $C$ as initial candidates.
Finally, we apply a non-maximum suppression algorithm to eliminate overlapping candidates, resulting in the final set of confirmed segments.

\begin{algorithm}[H]
\caption{The sliding window algorithm}\label{al:1}
\begin{algorithmic}[1]
\Statex \textbf{Input}: Preprocessed MEG and EEG recordings with $d$ channels and a total of $T_0$ time measurements; a segmentation time length $T$; a cutoff $C$.
\For{$t = T/2, \ldots, T_0 - T/2$}
\State Set the window between the $(t - T/2 + 1)$th and the $(t + T/2)$th time points.
\State Estimate the probability $\wh g_{(t)}$ and the weight $\wh\ba_{(l, t)}$ for Channel $l$.
\If{$\wh g_{(t)} > C$ and $\1\trans \wh\ba_{(l, t)} > (1.5 / d) (\sum_{l=1}^d \1\trans \wh\ba_{(l, t)})$}
\State Select the segment between the window as a candidate.
\EndIf
\State Remove duplicates by clustering the candidate segments based on time, utilizing the Density-Based Spatial Clustering of Applications with Noise (DBSCAN) method \citep{ester1996density}. 
\State Select the segment with the highest spike probability within each cluster as the final representation. 
\EndFor
\end{algorithmic}
\end{algorithm}

\vspace{-2mm}
We apply Algorithm~\ref{al:1} to the testing set of the UCSF MEG data and to the BTH EEG Data.
 Figure~\ref{fig:EEGanno}(a) presents the results of applying NDL to a continuous MEG recording from UCSF.
 The detection S1 accurately identifies the true spikes in the right temporal lobe, followed by an additional detection S2 shortly afterward.
 Although S2 appears outside the 0.25-second window according to the true spike labels, the detection remains valid since the true labels mark the onset of the spikes, and spikes may propagate to other channels over time.
 Moreover, our detection effectively highlights the channels where spikes occur, showing its accuracy in identifying spike-related activity across multiple channels.

\begin{figure}[htbp]

\begin{subfigure}{.48\textwidth}
\centering
\includegraphics[width=1.0\linewidth]{fig/MEG\_anno\_01.jpg}
\caption{MEG}
\vspace{-2mm}
\end{subfigure}
\begin{subfigure}{.51\textwidth}
\centering
\includegraphics[width=1.0\linewidth]{fig/EEG\_anno\_01.jpg}
\caption{EEG}
\vspace{-2mm}
\end{subfigure}
\caption{MEG and EEG recordings annotated by NDL using Algorithm~\ref{al:1}, with top channels highlighted in a dark color for each detection.
\label{fig:EEGanno}}
\vspace{-0mm}
\end{figure}

Figure~\ref{fig:EEGanno}(b) shows the results of applying NDL to the continuous EEG data from BTH.
 The detection S3 accurately identifies a true spike on the F7-avg channel.
 The detections S4 and S5 capture eye movements on the Fp1-avg and Fp2-avg channels, respectively.
 Eye movement artifacts typically appear simultaneously on these two channels since they are closest to the eyes.
 In practice, if both channels are identified as important, it is likely that the detection relates to eye movements rather than true spikes.
 Identification of key channels plays a crucial role in reducing false positives by helping to distinguish between eye movement artifacts and genuine spike activities.

\section{Conclusion and Discussion}
\label{sec:conc}

We have developed the NDL method as an interpretable alternative to traditional deep learning methods for spike detection.
This method can adapt to different channel types, making it a foundational model for neurophysiological data analysis.
We have established the model's identifiability and demonstrated the asymptotic consistency of its estimation procedure.
NDL has been applied to the publicly available TUH data, the private UCSF MEG data, and the private BTH EEG data.
Our analyses show that NDL outperforms existing methods in terms of sensitivity and specificity in spike detection.
Moreover, it effectively identifies channels containing relevant events.
Finally, we demonstrate that a model trained on the UCSF MEG data can be successfully applied to the BTH EEG data, highlighting NDL's capacity to integrate multiple datasets and serve as a foundation model that can be fine-tuned for different targets.

In this paper, our model is trained on the TUH data from 370 subjects and the UCSF MEG data from 277 subjects.
Due to the limited sample sizes, the model may not fully capture the variation in spike waveforms across different populations.
We are currently collaborating with clinicians to label more EEG and MEG data to improve the model's accuracy across a broader range of populations.
The tools developed in this paper will be used as a screening process to identify spike candidates, streamlining the labeling process in this larger study.

False discovery has long been a challenge in spike detection, and we acknowledge that our model trained on the limited samples does not fully address this issue.
However, identifying key relevant channels can help reduce false discoveries.
For instance, if the first two detected channels are those nearest to the eyes, the spike is more likely related to eye movements.
Additionally, training the model on larger datasets, which we are currently preparing, could significantly lower the false discovery rate.

\section*{Data Availability Statement}

The TUH data that support the findings of this study are available from the TUH EEG Corpus. Restrictions apply to the availability of these data, which were used under license for this study. Data are available at \url{https://www.isip.piconepress.com/projects/nedc/data/} with the permission of The Neural Engineering Data Consortium.

Due to the nature of the research, the UCSF MEG data and the BTH EEG data are not available due to commercial restrictions.

\if0\blind{
\section*{Acknowledgement}

The computations were performed using research computing facilities offered by Information Technology Services, the University of Hong Kong and by the Department of Radiology, the University of California, San Francisco.

\section*{Funding}
The work of Wei and Shen is partially supported by a HKSAR UGC CRF grant (C7162-20G). The work of Fei Jiang is supported by NIH grants K25AG071840 and R01NS132766.
}\fi

%%%%%%%%%%%%%%%%%%%%%%%%%%%%%%%%%%%%%%%%%%%%%%%%%%%%%%%%%%%%%%%%%%%%%%%%%%%%%%

% \newpage

% bibliographystyle{Chicago}
\bibliographystyle{agsm-asa}

\bibliography{Bibliography-MM-MC}

\def\spacingset#1{\renewcommand{\baselinestretch}%
{#1}\small\normalsize} \spacingset{1}

%%%%%%%%%%%%%%%%%%%%%%%%%%%%%%%%%%%%%%%%%%%%%%%%%%%%%%%%%%%%%%%%%%%%%%%%%%%%%%

%\if0\blind
%{
%  \title{\bf Supplementary Materials for ``Nested Deep Learning Model Towards a  Foundation Model for Brain Signal Data.''}
%  \author{Fangyi Wei \and
%        Jiajie Mo \and
%         Kai Zhang \and
% Haipeng Shen \and
%Srikantan Nagarajan \and
%    Fei Jiang
%    }
%    \date{}
%  \maketitle
%
%    \medskip
%
%} \fi
%
%\if1\blind
%{
%  \bigskip
%  \bigskip
%  \bigskip
%  \title{\bf Supplementary Materials for ``Nested Deep Learning Model Towards a  Foundation Model for Brain Signal Data.''}
%    \author{}
%    \date{}
%  
%  \maketitle
%  
%  \medskip
%} \fi

\allowdisplaybreaks

\newpage
 \begin{center}

   {\LARGE{\bf Supplementary Materials for ``Nested Deep Learning Model Towards a  Foundation Model for Brain Signal Data."}}\\
   
   \bigskip\bigskip
   
   {\large Fangyi Wei, \
        Jiajie Mo, \
         Kai Zhang, \
 Haipeng Shen, \ \\
Srikantan Nagarajan, \ and \
    Fei Jiang} \\
    
    \bigskip

 \end{center}

\setcounter{equation}{0}\renewcommand{\theequation}{S\arabic{equation}}
\setcounter{section}{0}\renewcommand{\thesection}{S\arabic{section}}
\setcounter{subsection}{0}\renewcommand{\thesubsection}{S\arabic{section}.\arabic{subsection}}
\setcounter{Pro}{0}\renewcommand{\thePro}{S\arabic{Pro}}
\setcounter{Th}{0}\renewcommand{\theTh}{S\arabic{Th}}
\setcounter{Lem}{0}\renewcommand{\theLem}{S\arabic{Lem}}
\setcounter{Rem}{0}\renewcommand{\theRem}{S\arabic{Rem}}
\setcounter{Cor}{0}\renewcommand{\theCor}{S\arabic{Cor}}
\setcounter{figure}{0}\renewcommand{\thefigure}{S\arabic{figure}}
\setcounter{table}{0}\renewcommand{\thetable}{S\arabic{table}}

\spacingset{1.9} % DON'T change the spacing!

\vspace{-10mm}

\section{Additional Notations}
We write $a \lesssim b$ if there exists a constant $C$ such that $a \leq Cb$ for all $n$.
Moreover, $a \asymp b$ means that $a \lesssim b$ and $b \lesssim a$.
%Let $\lfloor\beta\rfloor$ denote the largest integer strictly smaller than $\beta$, $\lceil\beta\rceil$ denote the smallest integer strictly larger than $\beta$.
Let $\mN(\delta, \mF, \|\cdot\|_\infty)$ be the covering number, that is, the minimal number of $\|\cdot\|_\infty$-balls with radius $\delta$ that covers $\mF$.
For a variable $X$, we define $\|X\|_{\psi_2} = \sup_{k \geq 1}
k^{-1/2} (E|X|^k)^{1/k}$ and define $\|X\|_{\psi_1} = \sup_{k \geq 1}
k^{-1} (E|X|^k)^{1/k}$.

\section{Useful Lemmas}
\begin{Lem}\label{lem:equ}
Let $\mF$ be a sub-field of the sigma-field generated by $X$.
Let $K_j(\mF)>0, j=1, \dots, 4$ be functions of random variables in $\mF$. 
Then, the following properties 1, 2,  and 3 are equivalent, and when
$E(X|\mF) = 0$, they are further equivalent to property 4. In addition,
$K_j(\mF)$ can be chosen to satisfy
$ 0<c<K_j(\mF) /K_k(\mF)<C<\infty$, 
for all $ k, j \in\{1,2,3, 4\}$, where $c, C$ are absolute constants.
\begin{enumerate}
\item Tail: There exists $K_1(\mF)$ such that
$\Pr(|X| >t |\mF) \leq \exp\{1 - t^2 /K_1^2
  (\mF)\}$; 
\item Moments: 
There exists $K_2(\mF)$ such that
$E(|X|^k|\mF)^{1/k} \leq K_2 (\mF) \sqrt{k}$,  for
  all $k \geq 1$; 
\item Super-exponential moment: 
 There exists $K_3(\mF)$ such that
$E[\exp\{X^2/K_3^2(\mF)\}|\mF]\leq
  e$; 
\item Let $E(X|\mF)=0$. There exists $K_4(\mF)$ such that 
$E\{\exp(t X)|\mF\}\leq
  \exp\{t^2 K_4^2(\mF)\}$. 
\end{enumerate}
\end{Lem}

\begin{Def}\label{def:guassian}
A random variable $X$ that satisfies one of the equivalent properties
in Lemma \ref{lem:equ} is named a conditional sub-Gaussian random
variable with respect to the sub-sigma field $\mF$. The
conditional 
sub-gaussian norm of $X$ with respect to $\mF$, denoted by
$\|X\|_{\psi_2 (\mF)}$, is defined as the smallest
$K_2(\mF)$ in property 2. That is, 
\bse
\|X\|_{\psi_2 (\mF)} = \sup_{k \geq 1 } k^{-1/2} E(|X|^{k}
|\mF)^{1/k}. 
\ese
\end{Def}

\begin{Lem}\label{lem:fdistance}
Assume $f \in \mG(q, \d, \t, \bb, K)$. Furthermore, let $\beta_{ i}^* \equiv \beta_{i}\prod_{l=i+1}^{q}
\{\min(\beta_{l}, 1) \}$ and $\phi_{n}\equiv \max_{i = 0, \ldots, q}
n^{-2\beta_{i}^*/(2 \beta^*_{i} + t_{i}) }$. Define a function space $\mF(L, (p_{wu})_{u = 0, \ldots, L+1}, s,
F)$ satisfying the following conditions:\\
(i) $F\geq \max(K, 1)$, \\
(ii) $\sum_{i=0}^q
\log_2\{\max(4t_i, 4\beta_i)\} \log_2 n\leq L\lesssim n\phi_n$,\\
(iii) $n\phi_n\lesssim \min_{u= 1, \ldots, L} p_{wu}$,\\(iv) $s
\asymp n\phi_n\log n$.\\
Then there exists a $f_0\in\mF(L, (p_{wu})_{u = 0, \ldots, L+1}, s, F) $ such that
\bse
\|f_0 - f\|_{\infty}^2 \leq C \max_{v = 0, \ldots, q} c^{-\frac{2
\beta_v^*}{t_v}} n^{-\frac{2 \beta_v^*}{2\beta_v^* + t_v}}. 
\ese
where $C, c$ are constants only depends on $q, \d, \t, \bb$. 
\end{Lem}
\noindent Proof: This lemma is a direct consequence of (26) in
\cite{schmidt2020nonparametric}.

\begin{Lem}\label{lem:coveringnumber}
For any $\delta>0$, we have
\bse
\log \mN\left\{\delta, \mF(L, \p_w, s, \infty), \|\cdot\|_{\infty}\right\} \leq (s + 1) \log\left\{ 2^{2 L + 5}
\delta^{-1} (L+1) p_{w0}^2 p_{w L+1}^2 s^{2L}\right\}. 
\ese
\end{Lem}
\noindent Proof: This lemma is a direct consequence of Lemma 4 in
\cite{schmidt2020nonparametric}.

\begin{Lem}{\label{lem:hoeffding}} Let $X_1, \ldots, X_N$ be
independent centered sub-Gaussian random variable, and let $K =
\max_i\|X_i\|_{\psi_2}$. Then for every $\a = (a_1, \ldots, a_N)\trans \in
\mathbb{R}^N$ and every $t \geq 0$, we have
\bse
\Pr\left(\bigg|\sum_{i=1}^N a_i X_i\bigg|\right) \leq
e \cdot \exp\left(-\frac{ct^2}{K^2 \|\a\|_2^2}\right), 
\ese
where $c >0$ is an absolute constant. 
\end{Lem}
\noindent Proof: This is a direct consequence of Proposition 5.10 in
\cite{vershynin2010introduction}.

\begin{Lem}\label{lem:Bernstein} Let $X_1, \ldots, X_N$ be
independent centered sub-exponential random variables, and $K =
\max_i\|X_i\|_{\psi_1}$. Then for every $\a = (a_1, \ldots,
a_N)\trans \in \mathbb{R}^N$ and every $t>0$, we have
\bse
\Pr\left(\bigg|\sum_{i=1}^N a_i X_i\bigg|\right) \leq
2\exp\left\{-c \min\left(\frac{t^2}{K^2 \|\a\|_2^2}, \frac{t}{K
\|\a\|_\infty}\right)\right\}, 
\ese
where $c>0$ is an absolute constant.
\end{Lem}

\section{Proof of Proposition \ref{pro:1}}\label{sec:pro:pro1}

\noindent Proof: For notation simplicity, we omit index $i$ in the following proofs. 
If $\ba(\X_{l},  \X)$ and the conditional density function $g\{\S(\ba)\}$  are not identifiable, then we
will have $\wt{g}$, $\wt{\ba}$ such that $(g,\ba)\neq (\wt{g},\wt{\ba})$, but 
\be\label{eq:gheq}
&&g\left\{\sum_{l=1}^d \X_{l}\ba(\X_{l}, \X)\trans +
\1_T\ba(\X_{l}, \X) \trans\Z_l  \1_p\trans \right \}\nonumber\\
&=& \wt
g\left\{\sum_{l=1}^{d} \X_{l}\wt{\ba}(\X_{l},
\X)\trans +  \1_{T}\wt{\ba}(\X_{l}, \X) \trans\Z_l \1_{p}\trans\right\}
\ee
for any given $\X_{l}, \Z_l$.
Now replacing $\Z_l$ by $\Z_l+ \xi I 
(\X_{l} = \X_u)\e_j$ in (\ref{eq:gheq}) where $\e_j$ is a length
 $p$ unit vector with the $j$th element
1, and  $I(\cdot)$ is an indicator function, 
Taking the derivative of both sides with respect to $\xi$ and
evaluate at $\xi=0$, denote $\M_{qt}$  be the ($q, t$)-th entry of matrix,
$\M$, then the left-hand side follows
\bse
&&\frac{\partial f\left\{\sum_{l=1}^d \X_{l}\ba(\X_{l}, \X)\trans +
\1_T\ba(\X_{l}, \X) \trans\{\Z_l  + \xi I 
(\X_{l} = \X_u)\e_j \} \1_p\trans \right \}}{\partial \xi}|_{\xi=0}\\
&=& \sum_{q=1}^p\sum_{t=1}^T\frac{\partial g\left\{\sum_{l=1}^d\X_{l}\ba(\X_{l}, \X)\trans +
\1_T\ba(\X_{l}, \X) \trans\Z \1_p\trans \right \} }{\partial \left\{\sum_{l=1}^d \X_{l}\ba(\X_{l}, \X)\trans +
\1_T\ba(\X_{l}, \X) \trans\Z  \1_p\trans\right\}_{qt} } \sum_{l=1}^d
\ba(\X_{l}, \X) \trans \e_j  I(\X_l=\X_u) \\
&=& \sum_{q=1}^p\sum_{t=1}^T\frac{\partial g\left\{\sum_{l=1}^d \X_{l}\ba(\X_{l}, \X)\trans +
\1_T\ba(\X_{l}, \X) \trans\Z \1_p\trans \right \} }{\partial
\left\{\sum_{l=1}^d \X_{l}\ba(\X_{l}, \X)\trans +\1_T\ba(\X_{l}, \X)
\trans\Z_l\1_p \trans\right\}_{qt}}  \alpha_j (\X_{u}, \X). 
\ese
The same follows for the right-hand side, and hence
we have 
\bse
\sum_{q=1}^p\sum_{t=1}^T \frac{\partial g\{\S(\ba)\}}{\partial S_{
qt}(\ba)}  \alpha_j(\X_{u}, \X) = \sum_{q=1}^p\sum_{t=1}^T
\frac{\partial \wt{g}\{\S(\wt{\ba})\}}{\partial S_{
qt}(\wt{\ba})}   \wt{\alpha}_j(\X_{u}, \X). 
\ese

Now summing across $\u$, noting that
$\partial g\{\S(\ba)\}/\partial S_{
qt}(\ba),   \partial \wt{g}\{\S(\wt{\ba})\}/{\partial S_{
qt}(\wt{\ba})}$ do not contain $\X_u$, and
$\sum_{u=1}^d{\alpha}_j(\X_{u},
\X) = \sum_{u=1}^d\wt \alpha_j(\X_{u},
\X)   = 1$ by our assumption,
we obtain
\bse
\sum_{q=1}^p\sum_{t=1}^T \frac{\partial g\{\S(\ba)\}}{\partial S_{
qt}(\ba)} = \sum_{q=1}^p\sum_{t=1}^T \frac{\partial g\{
\S(\wt{\ba})\}}{\partial S_{
qt}(\wt{\ba})}. 
\ese
And hence we have
\bse
{\alpha}_j(\X_{u}, \X) = \wt{\alpha}_j(\X_{u}, \X), j
= 1, \ldots, p, 
\ese
for any $\X$  and any $\u$.
Combined with (\ref{eq:gheq}), this  suggests
\bse
g\{\S(\ba)\}
&=& \wt
g\{\S(\ba)\}, 
\ese
for all $\X_{l}, \Z_{l}$, which implies $g = \wt{g}$. This contradicts with the assumption.  Therefore, $g$ and $\ba$ are identifiable.

%%%%%%%%%%%%%%%%%%%%%%%%%%%%%%%%%%%%%%%%%%%%%%%%%%%%%%%%%%%%%%%%
% \section{Theoretical Results}\label{sec:th}

% We derive the convergence rate and establish the asymptotic consistency of the estimators.

\section{Necessary Definitions and Conditions}
\label{sec:supp:cond}

% Below we first present some necessary definitions and conditions. 
% Let $g_0 \in \mF_g, \omega_{0k} \in \mF_{\omega_k}, k = 1, \ldots, p$ such that $g_0 \equiv \argmin_{g \in \mF_g} \|g - g^*\|_\infty$, and $\omega_{0k} \equiv \argmin_{\omega_k \in \mF_{\omega_k}}
% \|\omega_k - \omega^*_k\|_\infty$.
% where $g^*, \omega_k^*$ are the true functions in $\mG_g, \mG_{\omega_k}$, respectively.
% Let $\alpha_{0k}(\X_{il}, {\X_i}) =\exp\{\omega_{0k}(\X_{il})\}/\sum_{l=1}^d\exp\{\omega_{0k}(\X_{il})\}$ and $\ba_0 = \{\alpha_{0k}, k = 1, \ldots, p\}\trans$. Here, $g_0$ and $\omega_{0k}$ represent the functions within the neural network spaces $\mF_g$ and $\mF_{\omega_k}$ that are closest to $g^*$ and $\omega^*_k$, respectively.
% Let $\psi_g = g - g_0$ and $\psi_{\alpha_k} = \alpha_k - \alpha_{0k}$ for arbitrary functions $g\in \mF_g$ and $\alpha_k \in \mH_{\alpha_k}$, where
% \bse
% \mH_{k} \equiv \left\{\alpha_k =
%   \exp\{\omega_k(\X_{il})\}/\sum_{l=1}^d\exp\{\omega_k(\X_{il})\}:
%   \omega_k \in \mF_{\omega_k}, k = 1, \ldots, p\right\}.
% \ese
Let $\bpsi_{\ba} = \{ \psi_{\alpha_k}, k = 1, \ldots, p \} \trans$.
Furthermore, let $\ba^{(-k)} = (\alpha_{1}, \ldots, \alpha_{k-1}, \alpha_{k+1},
\ldots, \alpha_p)$ and 
\bse
&& \Omega(g_0, \ba_0, \psi_g, \bpsi_{\ba}, \epsilon_g^\dagger,
\beps_{\ba}^\dagger) \\
&=& \left\{ \frac{\partial^2 \mL(g_0  +
  \epsilon_g \psi_{g},\ba)}{\partial
  \epsilon_g^2 }+ 2\sum_{k = 1}^p\frac{\partial^2 \mL(g_0  +
  \epsilon_g \psi_{g}, \alpha_{0k} + \epsilon_{\alpha_k} \psi_{\alpha_k}, \ba^{(-k)})}{\partial
  \epsilon_g \partial \epsilon_{\alpha_k} } \right.\\
&&\left. + \sum_{k = 1}^p \sum_{j = 1}^p\frac{\partial^2 \mL(g_0, \alpha_{01}, \ldots, \alpha_{0l}+ \epsilon_{\alpha_l}
  \psi_{\alpha_l}, \ldots,  \alpha_{0u}+ \epsilon_{\alpha_u} \psi_{\alpha_u}, \ldots, \alpha_{0p})}{\partial \epsilon_{\alpha_j}\partial
  \epsilon_{\alpha_u} }\right\}\bigg|_{\epsilon_g = \epsilon^\dagger_g, \epsilon_{\alpha_{k}} =
  \epsilon_{\alpha_k}^\dagger, k = 1, \ldots, p}, 
\ese
where $\epsilon_g^{\dagger}, \epsilon_{\alpha_k}^{\dagger} \in (0, 1), k = 1, \ldots, p$ and $\beps_{\ba}^{\dagger} = (\epsilon_{\alpha_k}^{\dagger}, k = 1, \ldots, p)\trans$.
Define the rates
\be
\label{eq:rate}
\phi_{\omega_jn}\equiv
\max_{i = 0, \ldots, q_{\omega_j}}
n^{-\frac{2\beta_{\omega_ji}^*}{2 \beta^*_{\omega_j i} + t_{\omega_ji}}},
\phi_{gn}\equiv
\max_{i = 0, \ldots, q_g}
n^{-\frac{2\beta_{gi}^*}{2 \beta^*_{gi} + t_{gi}}}, 
\ee
where $\beta_{\omega_j i}^* \equiv \beta_{\omega_ji}
\prod_{l=i+1}^{q_{\omega_j}} \{\min(\beta_{\omega_jl}, 1) \}$ and $\beta_{g i}^* \equiv \beta_{g i}\prod_{l=i+1}^{q_{g}}
\{\min(\beta_{g l}, 1) \}$.
Then we present the necessary conditions to guarantee the convergence of the estimators.
\begin{enumerate}[label=(C\arabic*)]
\item \label{con:c1} %
Assume
$F_{\omega_j}\geq \max(K_{\omega_j}, 1), j = 1, \ldots, p$
and
$F_{g}\geq \max(K_{g}, 1)$.
\item \label{con:c2} %
Assume
$\sum_{i=0}^{q_{\omega_j}} \log_2\{\max(4t_{\omega_j i}, 4\beta_{\omega_ji})\}
\log_2 n \leq L_{\omega_j} \lesssim n\phi_{\omega_j n}, j = 1, \ldots, p$, \\
and
$\sum_{i=0}^{q_{g}} \log_2\{\max(4t_{g i}, 4\beta_{gi})\} \log_2 n
\leq L_g \lesssim n\phi_{g n}$.
\item \label{con:c3} %
Assume
$n\phi_{\omega_jn}\lesssim \min_{u= 1, \ldots, L} p_{\omega_j u}, j = 1, \ldots, p$
and
$n\phi_{g n}\lesssim \min_{u= 1, \ldots, L_g} p_{g u}$.
\item \label{con:c4} %
Assume
$s_{\omega_j} \asymp n\phi_{\omega_j n}\log n, j = 1, \ldots, p$
and
$s_g \asymp n\phi_{g n}\log n$.
And assume
$(s_g/L_g)^{L_g/2} \leq  M_L <\infty$.
\item \label{con:c5} %
% Assume
% $X_{ilt}, Z_{ilj}\in [- \max(K_{\omega_j}, K_{g}, j = 1, \ldots, p),
% \max(K_{\omega_j}, K_{g}, j = 1, \ldots, p)] \subseteq [-C_M, C_M]$. 
Assume
$|X_{ilt}|, |Z_{ilj}| \le
\max(K_{\omega_j}, K_{g}, j = 1, \ldots, p) \le C_M$. 
% \item \label{con:h} %
% Assume 
% $c_{\psi} <h''(\cdot ) <C_{\psi}$
% and
% $0 <|h'''(\cdot)| <\infty$,
% and
% $Y_i - h' \left[ g^*\left\{\S_i(\ba^*)\right\} \right]$
% is a mean zero sub-Gaussian random variable.
\item \label{con:RE} %
Assume that 
\bse
&& 0 <\alpha_{\min} E \left[\psi_g\left\{\S_i(\ba_0)\right\}^2 + \sum_{j=1}^p\sum_{l=1}^d T \psi_{\alpha_j}(\X_{il}, \X_i)^2\right]\\
&\leq& \inf_{\epsilon_g, \beps_{\ba}}E\{\Omega (g_0, \ba_0, \psi_g, \bpsi_{\ba}, \epsilon_g, \beps_{\ba})\} \leq  \sup_{\epsilon_g, \beps_{\ba}}E\{\Omega (g_0, \ba_0, \psi_g, \bpsi_{\ba}, \epsilon_g, \beps_{\ba})\}\\
&\leq& \alpha_{\max} E \left[\psi_g\left\{\S_i(\ba_0)\right\}^2 + \sum_{j=1}^p\sum_{l=1}^d T \psi_{\alpha_j}(\X_{il}, \X_i)^2\right] <\infty.
\ese
\end{enumerate}
Conditions \ref{con:c1}--\ref{con:c4} are standard assumptions to ensure that a deep learning network can approximate the true function with an error that vanishes as the sample size increases, as shown in Lemma \ref{lem:fdistance}.
% in the supplementary materials.
Condition \ref{con:c5} requires bounded predictors, which are easily met in real data.
% Conditions \ref{con:h}--\ref{con:RE}  are commonly assumed to ensure estimation consistency under the generalized linear model. 
Condition \ref{con:RE} is commonly assumed to ensure estimation consistency under the generalized linear model.

\section{Proof of Lemma \ref{lem:hh}}\label{sec:pro:lemm1}

\noindent Proof:
First note that 

\bse
\alpha_j(\x_l, \x) - \alpha_{0j}(\x_l, \x) &=&
\frac{\exp(\omega_j(\x_l))}{\sum_{l=1}^d\exp(\omega_j(\x_l))} -
\frac{\exp(\omega_{0j}(\x_l))}{\sum_{l=1}^d\exp(\omega_{0j}(\x_l))}\\
&=&
\frac{\exp(\omega^{\dagger}_j(\x_l))}{\sum_{l=1}^d\exp(\omega^\dagger_j(\x_l))}\left\{\left(1
  -\frac{\exp(\omega_{j}^{\dagger}(\x_l))}{\sum_{l=1}^d\exp(\omega_{j}^{\dagger} (\x_l))}\right
) \{\omega_j(\x_l) - \omega_{0j}(\x_l)\} \right.\\
&&\left. -\sum_{u\neq l} \frac{\exp(\omega_{j}^{\dagger}
    (\x_u))}{\sum_{l=1}^d\exp(\omega_{j}^{\dagger}
    (\x_l))}\{\omega_j(\x_u) - \omega_{0j}(\x_u)\}\right\},
\ese
where $\omega^\dagger_j$ is one the line between $\omega_0$ and
$\omega_j$, 
which implies 
\bse
\sup_{\X_{il}, \X_i} |\alpha_j(\X_{il}, \X_i) - \alpha_{0j}(\X_{il},\X_i) |\leq 2 \sup_{\X_{il}}
|\{\omega_j(\X_{il}) - \omega_{0j}(\X_{il})\}|.
\ese
Furthermore, 
\bse
\|\nabla  g_0\left \{\S_i(\ba_0) \right\} \|_2 \leq  \|\W_{gL} \W_{g L-1}, \ldots, \W_{g0}\|_2 \leq
 \sqrt{\prod_{l=1}^L s_{gl}},
 \ese
where $\W_{gl}, l = 0, \ldots, L$ are the weights for constructing
$g_0$, and $s_{gl}$ is the sparseness of $\W_{gl}$. By the
relationship between geometric and arithmetic means,  we have
$\prod_{l=1}^{L_g} s_{gl} \leq (\sum_{l=1}^{L_g}s_{gl} /L_g)^{L_g}$, which leads
 \be\label{eq:g0}
\|\nabla g_0\left \{\S_i(\ba_0) \right\} \|_2 \leq (s_g/L_g)^{L_g/2} 
  \leq M_L. 
  \ee
The last inequality holds by Condition \ref{con:c4}. 
Moreover, because $\nabla_{jt} g_0\left \{\S_i(\ba_0) \right\}  $ has at most $s_g$ nonzero
elements, 
 \bse
&& |\sum_{j=1}^p \sum_{t=1}^T \nabla_{jt} g_0\left \{\S_i(\ba_0)
\right\} \sum_{l=1}^d (X_{ilt} + Z_{ilj})\left \{
    \alpha_j^*(\X_{il}, \X_i)\trans  - \alpha_{j0}(\X_{il}, \X_i)\trans
  \right\} |\\
&\leq& 2C_M M_L\sqrt{d \min(s_g, Tp)}\sup_{\X_i, j = 1, \ldots, p} |\alpha^*_j(\X_{il}, \X_i) -\alpha_{j0}(\X_{il},
\X_i) |. 
\ese
Therefore, by Lemma \ref{lem:fdistance}, we have 
\bse
&& |h'\left[g^*\left \{\S_i(\ba^*) \right\}\right] -h'\left[g_0\left \{\S_i(\ba_0) \right\}\right]| \\
&=& \bigg| h''\left[g^\dagger \left \{\S_i(\ba^\dagger)  \right\}\right] \left[g^*\left \{\S_i(\ba_0)  \right\}- g_0\left \{\S_i(\ba_0) \right\}\right] + h''\left[g^\dagger \left \{\S_i(\ba^\dagger)  \right\}\right] \\
&& \times \sum_{j=1}^p\sum_{t=1}^T
\nabla_j g^\dagger \left \{\S_i(\ba^\dagger) \right\} \sum_{l=1}^d (X_{ilt} + Z_{ilj})\left \{
    \alpha_j^*(\X_{il}, \X_i)\trans  - \alpha_{j0}(\X_{il}, \X_i)\trans
  \right\} \bigg|\\
&\leq& \left\{C_g \max_{u = 0, \ldots, q_g} c_g^{-\frac{
\beta_{gu}^*}{t_{gu}}} n^{-\frac{ \beta_{gu}^*}{2\beta_{gu}^* +
    t_{gu}}} +2C_M M_L\sqrt{d \min(s_g, Tp)} \right. \\
&& \left. \sup_{j=1, \ldots, p} C_{\omega_j} \max_{u = 0, \ldots, q_{\omega_j}} c_{\omega_j}^{-\frac{
\beta_{\omega_j u}^*}{t_{\omega_j u}}} n^{-\frac{\beta_{\omega_j
u}^*}{2\beta_{\omega_j u}^* + t_{\omega_j u}}}\right\}. 
\ese

\section{Useful Lemmas for Theorem \ref{th:1}}

\begin{Lem}\label{lem:firstorder}
Let $\psi_{\wh g}= \wh{g} - g_0$ and $\psi_{\wh \alpha_j} = \wh{\alpha}_j - \alpha_{0j}$. 
Assume $\wh{\alpha}_j, \alpha_{0j} \in \mH_{j}$, $\omega^*_j \in \mG_{\omega_j}$, and
$\wh{g}, g_0 \in \mF_g, $ and $g^* \in \mG_{g}$ and Conditions
% \ref{con:c1}--\ref{con:h} hold,  and $T, p \leq s_g$.  
\ref{con:c1}--\ref{con:c5} hold,  and $T, p \leq s_g$.  
Then there are positive constants $C_g,
C_{\omega_j}, c_g, c_{\omega_j}$,  $\C_{\omega} =(C_{\omega_j}, j = 1,
\ldots, p)\trans, \c_{\omega} =(c_{\omega_j}, j = 1,
\ldots, p)\trans$  such that  \bse
 && \bigg |n^{-1}
\sumi\left\{\mystrut \left( Y_i -h'\left[g_0\left \{\S_i(\ba_0) \right\}\right] \right) \right.\nonumber\\
&&\left.\sum_{j=1}^p \sum_{t=1}^T\nabla_{jt}  g_0\left \{\S_i(\ba_0)
  \right\} \sum_{l =1}^d (X_{ilt} + Z_{ilj}) \psi_{\wh{h}_j}(\X_{il}, \X_i)\right\}\bigg|\nonumber\\
&\leq& 4C_M M_L \eta(C_g, \C_{\omega}, c_g, \c_{\omega}) \left(\left[ dp E\left\{\sum_{j=1}^p\sum_{l =1}^d
  \{\sqrt{T} \psi_{\wh{\alpha}_j}(\X_{il},
  \X_i)\}^2\right\}\right]^{1/2} \right.\\
&&\left.\mystrut+ \min(s_g, Tp)^{1/2}
  \{\log(n)/(cn)\}^{1/4}\right)
\ese
with probability greater than $1 - 6n^{-1}$. 
And there are positive constants $B_g,
B_{\omega_j}, b_g, b_{\omega_j}$,  $\B_{\omega} =(B_{\omega_j}, j = 1,
\ldots, p)\trans, \b_{\omega} =(b_{\omega_j}, j = 1,
\ldots, p)\trans$  that 
\bse
&& \bigg |n^{-1}
\sumi\left\{\left( Y_i - h' \left[g_0\left \{\S_i(\ba_0) \right\}\right] \right) \psi_{\wh{g}} \left\{\S_i(\ba_0)\right\}\right\} \bigg|\\
 &\leq& 4C_M M_L \eta(B_g, B_g, \B_{\omega}, \b_{\omega}) \left[\left(E
\left[ \psi_{\wh{g}} \left\{\S_i(\ba_0) \right\}^2 \right] \right)^{1/2}+ \{\log(n)/(cn)\}^{1/4}\right]. 
\ese
with probability greater than $1- 6n^{-1}$. 
\end{Lem}

\noindent Proof:
Because $Y_i - h'[
g^*\left\{\S_i(\ba^*)\right\}]$  is a sub-Gaussian random variable, by
Lemma \ref{lem:hh} that $|h'\left[g^*\left \{\S_i(\ba^*)
  \right\}\right] -h'\left[g_0\left \{\S_i(\ba_0) \right\}\right]|\to
0$ almost surely.
% , we have $Y_i - h'[g_0\left\{\S_i(\ba_0)\right\}]$ is a sub-Gaussian random variable.
We denote
\bse
\tau_i = \sum_{j=1}^p\sum_{t=1}^T\nabla_{jt} g_0\left\{\S_i(\ba_0)\right\} \sum_{l =1}^d (X_{ilt} +Z_{ilj})\psi_{\wh{\alpha}_j}(\X_{il}, \X_i)
\ese
in the following proof for convenience.
Therefore
using the concentration inequality in Lemma \ref{lem:hoeffding},  we have
% \be\label{eq:problessY}
% &&\Pr\left[\bigg|n^{-1}
% \sumi\left\{\left( Y_i - h'\left[g_0\left \{\S_i(\ba_0)\right\}\right] \right)
% - E \left( Y_i - h'\left[g_0\left \{\S_i(\ba_0) \right\}\right] |\mathcal{X}_n\right) \right\} \right.\nonumber\\
% &&\left.\times \sum_{j=1}^p\sum_{t=1}^T\nabla_{jt} g_0\left
%       \{\S_i(\ba_0)\right\} \sum_{l =1}^d (X_{ilt} +Z_{ilj})\psi_{\wh{\alpha}_j}(\X_{il}, \X_i)\bigg|
%   \geq t  \bigg|\mathcal{X}_n\right]\\
% & \leq&  3\exp\left(-\frac{n c t^2}{ 16 n^{-1}\sumi \left[\sum_{j=1}^p\sum_{t=1}^T\nabla_{jt} g_0\left
%       \{\S_i(\ba_0)\right\} \sum_{l =1}^d (X_{ilt} +Z_{ilj})\psi_{\wh{\alpha}_j}(\X_{il}, \X_i)\right]^2}\right). \nonumber
% \ee
\be\label{eq:problessY}
&&\Pr\left[\bigg|n^{-1}
\sumi\left\{\left( Y_i - h'\left[g_0\left \{\S_i(\ba_0)\right\}\right] \right)
- E \left( Y_i - h'\left[g_0\left \{\S_i(\ba_0) \right\}\right] |\mathcal{X}_n\right) \right\}
 \tau_i \bigg|
  \geq t  \bigg|\mathcal{X}_n\right] \nonumber\\
& \leq&  3\exp\left[ -{n c t^2} \bigg/ \left( 16 n^{-1}\sumi \tau_i^2 \right) \right]. 
\ee
Let $$t = \sqrt{C_0 \log(n)/(cn)} \left(n^{-1}\sumi \left[\sum_{j=1}^p\sum_{t=1}^T\nabla_{jt} g_0\left
      \{\S_i(\ba_0)\right\} \sum_{l =1}^d (X_{ilt} +Z_{ilj})\psi_{\wh{\alpha}_j}(\X_{il}, \X_i)\right]^2\right)^{1/2}, $$
where $C_0= 16$, we
obtain
% \be\label{eq:R1}
% && \left|n^{-1}
% \sumi\left\{\left( Y_i - h'\left[g_0\left \{\S_i(\ba_0)\right\}\right] \right)  - E \left( Y_i - h'\left[g_0\left \{\S_i(\ba_0) \right\}\right]|\mathcal{X}_n \right) \right\}\right. \nonumber\\
% &&\left.\sum_{j=1}^p\sum_{t=1}^T\nabla_{jt} g_0\left \{\S_i(\ba_0) \right\} \sum_{l =1}^d
%   (X_{ilt} + Z_{ilj})\psi_{\wh{\alpha}_j}(\X_{il}, \X_i)\right| \\
% & \leq &  \sqrt{C_0 \log(n)/(cn)} \left(n^{-1}\sumi \left[\sum_{j=1}^p\sum_{t=1}^T\nabla_{jt} g_0\left
%       \{\S_i(\ba_0)\right\} \sum_{l =1}^d (X_{ilt} +Z_{ilj})\psi_{\wh{\alpha}_j}(\X_{il}, \X_i)\right]^2\right)^{1/2} \nonumber
% \ee
\be\label{eq:R1}
&& \left|n^{-1}
\sumi\left\{\left( Y_i - h'\left[g_0\left \{\S_i(\ba_0)\right\}\right] \right)  - E \left( Y_i - h'\left[g_0\left \{\S_i(\ba_0) \right\}\right]|\mathcal{X}_n \right) \right\}
\tau_i \right| \\
& \leq &  \sqrt{C_0 \log(n)/(cn)} \left(n^{-1}\sumi \tau_i ^2\right)^{1/2} \nonumber
\ee
with probability greater than $1 - 3n^{-1}$. 

Furthermore,  Lemma \ref{lem:hh} leads to 
% \bse 
% && \bigg|n^{-1}\sumi E\left\{\left( Y_i -h'\left[g_0\left \{\S_i(\ba_0) \right\}\right] \right) |\mathcal{X}_n\right\} \\
% && \sum_{j=1}^p\sum_{t=1}^T \nabla_{jt} g_0\left \{\S_i(\ba_0)
% \right\} \sum_{l =1}^d (X_{ilt} +Z_{ilj})\psi_{\wh{\alpha}_j}(\X_{il}, \X_i)\bigg|\\
% &=& \bigg|n^{-1}\sumi\left\{\mystrut \left( h'\left[g^*\left \{\S_i(\ba^*) \right\}\right] -h'\left[g_0\left \{\S_i(\ba_0) \right\}\right] \right) \right.\\
% &&\left. \sum_{j=1}^p\sum_{t=1}^T\nabla_{jt}  g_0\left \{\S_i(\ba_0) \right\}
%   \sum_{l =1}^d (X_{ilt} +Z_{ilj}) \psi_{\wh{\alpha}_j}(\X_{il}, \X_i)\right\}\bigg|\\
% &\leq& \left\{C_g \max_{u = 0, \ldots, q_g} c_g^{-\frac{
%     \beta_{gu}^*}{t_{gu}}} n^{-\frac{ \beta_{gu}^*}{2\beta_{gu}^* +
%     t_{gu}}} +2 C_MM_L\sqrt{d \min(s_g, Tp)} \right. \\
%     && \left. \sup_{j=1, \ldots, p}C_{\omega_j} \max_{u = 0, \ldots, q_{\omega_j}} c_{\omega_j}^{-\frac{
%     \beta_{\omega_j u}^*}{t_{\omega_j u}}} n^{-\frac{\beta_{\omega_j
%       u}^*}{2\beta_{\omega_j u}^* + t_{\omega_j u}}}\right\}\\
% && \left( n^{-1}\sumi \left[\sum_{j=1}^p\sum_{t=1}^T\nabla_{jt}
%     g_0\left \{\S_i(\ba_0) \right\} \sum_{l =1}^d
%   (X_{ilt} + Z_{ilt})\psi_{\wh{\alpha}_j}(\X_{il}, \X_i)\right]^2\right)^{1/2}.
% \ese
\bse 
&& \bigg|n^{-1}\sumi E\left\{\left( Y_i -h'\left[g_0\left \{\S_i(\ba_0) \right\}\right] \right) |\mathcal{X}_n\right\} 
\tau_i \bigg|\\
&=& \bigg|n^{-1}\sumi\left\{\mystrut \left( h'\left[g^*\left \{\S_i(\ba^*) \right\}\right] -h'\left[g_0\left \{\S_i(\ba_0) \right\}\right] \right)
\tau_i \right\}\bigg|\\
&\leq& \left\{C_g \max_{u = 0, \ldots, q_g} c_g^{-\frac{
    \beta_{gu}^*}{t_{gu}}} n^{-\frac{ \beta_{gu}^*}{2\beta_{gu}^* +
    t_{gu}}} +2 C_MM_L\sqrt{d \min(s_g, Tp)} \right. \\
    && \left. \sup_{j=1, \ldots, p}C_{\omega_j} \max_{u = 0, \ldots, q_{\omega_j}} c_{\omega_j}^{-\frac{
    \beta_{\omega_j u}^*}{t_{\omega_j u}}} n^{-\frac{\beta_{\omega_j
      u}^*}{2\beta_{\omega_j u}^* + t_{\omega_j u}}}\right\}
      \left( n^{-1}\sumi \tau_i ^2\right)^{1/2}.
\ese
Combine with (\ref{eq:R1}), we obtain there are positive constants $C_g,
C_{\omega_j}, c_g, c_{\omega_j}$ such that 
% \be\label{eq:yH}
%  && \bigg |n^{-1}
% \sumi\left\{\mystrut \left( Y_i -h'\left[g_0\left \{\S_i(\ba_0)\right\}\right] \right) \right.\nonumber\\
% &&\left. \sum_{j=1}^p\sum_{t=1}^T\nabla_{jt} g_0\left
%     \{\S_i(\ba_0)\right\} \sum_{l =1}^d (X_{ilt} +Z_{ilj})\psi_{\wh{\alpha}_j}(\X_{il}, \X_i)\right\}\bigg| \nonumber \\
% &\leq& \eta(C_g, \C_{\omega}, c_g, \c_{\omega}) \left(n^{-1}\sumi \left[\sum_{j=1}^p \sum_{t=1}^T\nabla_{jt} g_0\left \{\S_i(\ba_0)\right\} \right.\right. \nonumber \\
% && \left.\left. \times \sum_{l =1}^d
%   (X_{ilt} +Z_{ilj})\psi_{\wh{\alpha}_j}(\X_{il}, \X_i) \right]^2\right)^{1/2},
% \ee
\be\label{eq:yH}
 && \bigg |n^{-1}
\sumi\left\{\mystrut \left( Y_i -h'\left[g_0\left \{\S_i(\ba_0)\right\}\right] \right) 
\tau_i \right\}\bigg| \nonumber \\
&\leq& \eta(C_g, \C_{\omega}, c_g, \c_{\omega}) \left(n^{-1}\sumi \tau_i ^2\right)^{1/2},
\ee
with probability greater than $1 - 3n^{-1}$.
Additionally, because $\nabla  g_0\left \{\S_i(\ba_0)\right\}$ is $s_g$ sparse and $\sum_{l=1}^d\wh{\alpha}_j(\X_{il}, \X_i) =
\sum_{l=1}^d{\alpha}_{0j}(\X_{il}, \X_i)  = 1$, we have 
% \bse
% | \sum_{j=1}^p \sum_{t=1}^T \nabla_{jt} g_0\left \{\S_i(\ba_0) \right\} \sum_{l =1}^d
%   (X_{ilt} +Z_{ilj})\psi_{\wh{\alpha}_j}(\X_{il}, \X_i)| \leq 
% 4C_M M_L \sqrt{\min(s_g, Tp)}. 
% \ese
\bse
| \tau_i | \leq 
4C_M M_L \sqrt{\min(s_g, Tp)}. 
\ese
% Therefore  $\left\{\sum_{j=1}^p\sum_{t=1}^T \nabla_{jt} g_0\left \{\S_i(\ba_0)\right\} \sum_{l =1}^d (X_{ilt} + Z_{ilj})\psi_{\wh{\alpha}_j}(\X_{il}, \X_i)\right\}^2/\min(s_g, Tp)$ is sub-Gaussian variable, and hence
Therefore $\tau_i ^2/\min(s_g, Tp)$ is sub-Gaussian variable, and hence
% \bse
% &&   \Pr\left\{\bigg|n^{-1}\min(s_g, Tp)^{-1}\sumi \left[\sum_{j=1}^p\sum_{t=1}^T\nabla_{jt} g_0\left \{\S_i(\ba_0) \right\} \sum_{l =1}^d
%   (X_{il} + Z_{ilj})\psi_{\wh{\alpha}_j}(\X_{il}, \X_i) \right]^2 \right.\\
% && \left.- \min(s_g, Tp)^{-1} E\left(\left[\sum_{j=1}^p\sum_{t=1}^T\nabla_{jt} g_0\left \{\S_i(\ba_0)\right\} \sum_{l =1}^d
%   (X_{il} + Z_{ilj})\psi_{\wh{\alpha}_j}(\X_{il}, \X_i) \right]^2\right)\bigg|>t
% \right\} \nonumber\\
% &\leq& 3 \exp\left\{-c nt^2/(256 C_M^4 M_L^4)\right). 
% \ese
\bse
&&   \Pr\left\{\bigg|n^{-1}\min(s_g, Tp)^{-1}\sumi \tau_i ^2 
- \min(s_g, Tp)^{-1} E\left(\tau_i^2\right)\bigg|>t
\right\} \nonumber\\
&\leq& 3 \exp\left\{-c nt^2/(256 C_M^4 M_L^4)\right). 
\ese
Let $t = \sqrt{256 M_L^4 C_M^4 \log(n)/(cn)}$, we have
% \be\label{eq:sumgeg}
%  &&   \bigg|n^{-1}\sumi \left\{\sum_{j=1}^p\sum_{t=1}^T\nabla_{jt} g_0\left \{\S_i(\ba_0)\right\} \sum_{l =1}^d
%   (X_{ilt} + Z_{ilj}) \psi_{\wh{\alpha}_j}(\X_{il}, \X_i) \right\}^2 \nonumber\\
% && - E\left(\left\{\sum_{j=1}^p\sum_{t=1}^T\nabla_{jt} g_0\left \{\S_i(\ba_0)\right\} \sum_{l =1}^d
%   (X_{il} + Z_{ilj})\psi_{\wh{\alpha}_j}(\X_{il}, \X_i) \right\}^2\right)\bigg|\nonumber\\
% &\leq&
% 16 \min(s_g, Tp) C_M^2 M_L^2\sqrt{ \log(n)/(cn)}, 
% \ee
\be\label{eq:sumgeg}
 \bigg|n^{-1}\sumi \tau_i ^2 - E\left(\tau_i ^2\right)\bigg|
\leq
16 \min(s_g, Tp) C_M^2 M_L^2\sqrt{ \log(n)/(cn)}, 
\ee
with probability greater than $1 -3/n$. 
Next
% \bse
% && E\left(\left[\sum_{j=1}^p \sum_{t=1}^T\nabla_{jt} g_0\left \{\S_i(\ba_0)\right\} \sum_{l =1}^d
%   (X_{ilt} + Z_{ilj})\psi_{\wh{\alpha}_j}(\X_{il}, \X_i) \right]^2\right)\\
% &\leq& E\left( \left[\bigg|\sup_{l = 1, \ldots, d, j = 1, \ldots, p}
%    \sum_{t=1}^T \nabla_{jt} g_0\left \{\S_i(\ba_0) \right\}  (X_{ilt} + Z_{ilj}) /\sqrt{T}\bigg| \right.\right. \\
%    && \left.\left. \times \sum_{j=1}^p \sum_{l =1}^d
%   \bigg|\sqrt{T} \psi_{\wh{\alpha}_j}(\X_{il}, \X_i)\bigg| \right]^2\right)\\
% &\leq& 4C^2_M M^2_L  E\left\{\left(\sqrt{dp} \left[\sum_{j=1}^p\sum_{l =1}^d
%  \left\{\sqrt{T} \psi_{\wh{\alpha}_j}(\X_{il}, \X_i)\right\}^2\right]^{1/2}
% \right)^2 \right\}\\
% &=& 4C^2_M M^2_L d p  E\left[\sum_{j=1}^p\sum_{l =1}^d
%   \left\{\sqrt{T} \psi_{\wh{\alpha}_j}(\X_{il}, \X_i)\right\}^2\right], 
% \ese
\bse
 E\left(\tau_i^2\right)
&\leq& E\left( \left[\bigg|\sup_{l = 1, \ldots, d, j = 1, \ldots, p}
   \sum_{t=1}^T \nabla_{jt} g_0\left \{\S_i(\ba_0) \right\}  (X_{ilt} + Z_{ilj}) /\sqrt{T}\bigg| \right.\right. \\
   && \left.\left. \times \sum_{j=1}^p \sum_{l =1}^d
  \bigg|\sqrt{T} \psi_{\wh{\alpha}_j}(\X_{il}, \X_i)\bigg| \right]^2\right)\\
&\leq& 4C^2_M M^2_L  E\left\{\left(\sqrt{dp} \left[\sum_{j=1}^p\sum_{l =1}^d
 \left\{\sqrt{T} \psi_{\wh{\alpha}_j}(\X_{il}, \X_i)\right\}^2\right]^{1/2}
\right)^2 \right\}\\
&=& 4C^2_M M^2_L d p  E\left[\sum_{j=1}^p\sum_{l =1}^d
  \left\{\sqrt{T} \psi_{\wh{\alpha}_j}(\X_{il}, \X_i)\right\}^2\right], 
\ese
for a constant $C_M$, where the second inequality holds because
\bse
&& \bigg|\sup_{l = 1, \ldots, d, j = 1, \ldots, p}
   \sum_{t=1}^T \nabla_{jt} g_0\left \{\S_i(\ba_0) \right\}  (X_{ilt} +
   Z_{ilj}) \bigg|\\
   &\leq& 2C_M\sup_{l = 1, \ldots, d, j = 1, \ldots, p}
   \sum_{t=1}^T |\nabla_{jt} g_0\left \{\S_i(\ba_0) \right\} |\\
   &\leq& 2C_M\sqrt{T} \sup_{l = 1, \ldots, d, j = 1, \ldots, p}
   \left[\sum_{t=1}^T |\nabla_{jt} g_0\left \{\S_i(\ba_0) \right\} |\right]^{1/2}\\
   &\leq& 2C_M\sqrt{T} M_L, 
   \ese
   by (\ref{eq:g0}). 
Plugging the above inequality and (\ref{eq:sumgeg}) to (\ref{eq:yH}),
we obtain
% \bse
%  && \bigg |n^{-1}
% \sumi\left\{\left( Y_i -h'\left[g_0\left \{\S_i(\ba_0) \right\}\right] \right) \right.\nonumber\\
% &&\left.  \sum_{j=1}^p\sum_{t=1}^T\nabla_j  g_0\left \{\S_i(\ba_0)  \right\}
%   \sum_{l =1}^d (X_{ilt} +Z_{ilj})\psi_{\wh{\alpha}_j}(\X_{il}, \X_i)\right\}\bigg|\nonumber\\
% &\leq& \eta(C_g, \C_{\omega}, c_g, \c_{\omega}) \left(2C_M M_L\left[ dp E\left\{\sum_{j=1}^p\sum_{l =1}^d
%   \{\sqrt{T} \psi_{\wh{\alpha}_j}(\X_{il},
%   \X_i)\}^2\right\}\right]^{1/2} \right.\\
% &&\left.\mystrut+ 4 C_M M_L \min(s_g, Tp)^{1/2}
%   \{\log(n)/(cn)\}^{1/4}\right)\\
% &\leq&  4C_M M_L \eta(C_g, \C_{\omega}, c_g, \c_{\omega}) \left(\left[ dp E\left\{\sum_{j=1}^p\sum_{l =1}^d
%   \{\sqrt{T} \psi_{\wh{\alpha}_j}(\X_{il},
%   \X_i)\}^2\right\}\right]^{1/2} \right.\\
% &&\left.\mystrut+ \min(s_g, Tp)^{1/2}
%   \{\log(n)/(cn)\}^{1/4}\right)
% \ese
\bse
 && \bigg |n^{-1}
\sumi\left\{\left( Y_i -h'\left[g_0\left \{\S_i(\ba_0) \right\}\right] \right) \tau_i \right\}\bigg|\nonumber\\
&\leq& \eta(C_g, \C_{\omega}, c_g, \c_{\omega}) \left(2C_M M_L\left[ dp E\left\{\sum_{j=1}^p\sum_{l =1}^d
  \{\sqrt{T} \psi_{\wh{\alpha}_j}(\X_{il},
  \X_i)\}^2\right\}\right]^{1/2} \right.\\
&&\left.\mystrut+ 4 C_M M_L \min(s_g, Tp)^{1/2}
  \{\log(n)/(cn)\}^{1/4}\right)\\
&\leq&  4C_M M_L \eta(C_g, \C_{\omega}, c_g, \c_{\omega}) \left(\left[ dp E\left\{\sum_{j=1}^p\sum_{l =1}^d
  \{\sqrt{T} \psi_{\wh{\alpha}_j}(\X_{il},
  \X_i)\}^2\right\}\right]^{1/2} \right.\\
&&\left.\mystrut+ \min(s_g, Tp)^{1/2}
  \{\log(n)/(cn)\}^{1/4}\right)
\ese
with probability greater than $1 - 6n^{-1}$. 
Using the same arguments and realize that we obtain there are positive constants $B_g,
B_{\omega_j}, b_g, b_{\omega_j}$  and realize that $\|\psi_{\wh{g}}\|_2\leq 2
  C_MM_L$, we have 
\bse
&& \bigg |n^{-1}
\sumi\left\{\left( Y_i - h' \left[g_0\left \{\S_i(\ba_0)\right\}\right] \right) \psi_{\wh{g}} \left\{\S_i(\ba_0)\right\}\right\} \bigg|\\
 &\leq&  4C_M M_L \eta(B_g, \B_{\omega},b_g, \B_{\omega})\left[\left(E
\left[ \psi_{\wh{g}} \left\{\S_i(\ba_0)\right\}^2 \right]\right)^{1/2} +\{\log(n)/(cn)\}^{1/4}\right]. 
\ese
with probability greater than $1- 6n^{-1}$. 
This proves the result.

\begin{Lem}\label{lem:secondorder}
Assume $\wh{\alpha}_j, \alpha_{0j} \in \mH_{j}$, $\omega^*_j \in \mG_{\omega_j}$, and
 $\wh{g}, g_0 \in \mF_g $, $g^* \in \mG_{g}$.  Furthermore,
assume Condition \ref{con:c1}--\ref{con:RE} hold, and $d /n \to 0$ and $T/n\to
0$ and $p < \infty$. Then for given $\epsilon_g^\dagger, \epsilon_{\alpha_j}^\dagger\in (0, 1), \beps^\dagger_{\ba}
 = \{\epsilon_{\alpha_j}^\dagger, j = 1, \ldots, p\}\trans$, we have
\bse
&& \Omega (g_0, \ba_0, \psi_{\wh g}, \bpsi_{\wh{\ba}}, \epsilon_g^\dagger,
\beps_{\ba}^\dagger) \geq  \alpha_{\min} E \left[\psi_{\wh{g}}\left\{\S_i(\ba_0)\right\}^2 + \sum_{j=1}^p\sum_{l=1}^d T
   \psi_{\wh{\alpha}_j}(\X_{il}, \X_i)^2\right]\\
 && -   2Tp C_1 \log^2 (n) \left\{\phi_{gn} L_g  + \sum_{j=1}^p \phi_{\omega_j n} L_{\omega_j} \right\} - C_H T  p/n
\ese
with probability greater than $1 -  2\exp \left[- n C_1\log^2 (n) \left\{\phi_{gn} L_g  + \sum_{j=1}^p \phi_{\omega_j n} L_{\omega_j} \right\} \right]$
for some constants
$C_H, C_1>0$. 
  \end{Lem}
  \noindent Proof:
Let $\psi_{\wh{g}} = \wh{g} - g_0$, $\psi_{\wh{\alpha}_k} = \wh{\alpha}_k -
\alpha_0$. First, because $\wh{g}, g_0, \wh{\alpha}_k, \alpha_0$ are bounded
functions, $\bigtriangledown g$ is bounded in the sense of
\ref{eq:g0}, and $Y_i$ is bounded,  and $\|\X_{il}\|_2\leq C_M
\sqrt{T}$ by Condition \ref{con:c5},  it is easy to see that every summand in
$ {Tp}^{-1}\Omega(g_0, \ba_0, \psi_{ g}, \bpsi_{\ba}, \epsilon_g^\dagger,
\beps_{\ba}^\dagger)$ is a product of bounded sub-Gaussian variable, and hence
sub-exponential. Therefore, by Lemma \ref{lem:Bernstein}, for arbitrary $g, \ba$, there is a constant $c_H>0$ such that 
\bse
&& \Pr\left\{|(Tp)^{-1}\Omega (g_0, \ba_0, \psi_{ g}, \bpsi_{\ba}, \epsilon_g^\dagger,
\beps_{\ba}^\dagger) -(Tp)^{-1}  E\{\Omega (g_0, \ba_0, \psi_{ g}, \bpsi_{\ba}, \epsilon_g^\dagger,
\beps_{\ba}^\dagger) \} | > t\right\} \\
&\leq& 3 \exp\{- c_H n\min(t^2, t)\}. 
\ese
Let $\mS_g(\delta)$ be the $\delta$ covering of $\mF_g$ and
$\mS_{\omega_k}(\delta)$ be the $\delta$ covering of $\mF_{\omega_k}$,
then for some constant $C_H>0$, we have 
\bse
&& \Pr\left\{\sup_{g \in \mF_{g}, \alpha_k\in \mH_k}|\Omega (g_0, \ba_0, \psi_{ g}, \bpsi_{\ba}, \epsilon_g^\dagger,
\beps_{\ba}^\dagger) - E\{\Omega (g_0, \ba_0, \psi_{ g}, \bpsi_{\ba}, \epsilon_g^\dagger,
\beps_{\ba}^\dagger) \} |  >Tp t + C_H T  p\delta\right\}\\
&\leq &  \Pr\left\{\max_{g \in \mS_{g}, \omega_k\in \mS_{\omega_k}}|\Omega (g_0, \ba_0, \psi_{g}, \bpsi_{\ba}, \epsilon_g^\dagger,
\beps_{\ba}^\dagger) - E\{\Omega (g_0, \ba_0, \psi_{ g}, \bpsi_{\ba}, \epsilon_g^\dagger,
\beps_{\ba}^\dagger) \} |  >Tp t \right\}\\
&\leq &2 \exp\left[- c_H n\min(t^2, t) + \log \mN\left\{\delta, \mF_g,
  \|\cdot\|_{\infty}\right\}  + \sum_{k=1}^p\log \mN\left\{\delta,
  \mF_{\omega_k}, \|\cdot\|_{\infty}\right\}\right]\\
&\leq& 2 \exp\left[- c_H n\min(t^2, t) + (s_g + 1) \log\left\{ 2^{2 L_g + 5}
    \delta^{-1} (L+1) p_{g0}^2 p_{g L_g+1}^2 s_g^{2L_g} \right\}\right.\\
&&\left.+
    \sum_{k=1}^p (s_{\omega_k} + 1) \log\left\{ 2^{2 L_{\omega_k} + 5}
    \delta^{-1} (L_g \omega_k+1) p_{\omega_k 0}^2 p_{\omega_k
      L_{\omega_k}+1}^2 s_{\omega_k}^{2L_{\omega_k}}\right\}\right]\\
&\leq& 2 \exp\left[ \mystrut - c_H n\min(t^2, t) + (s_g + 1)  L_g \log\left\{ 
    2s_g 
    \delta^{-1} (L_g+1) d^2p^2 s_g^{2L_g} \right\}\right.\\
&&\left.+
    \sum_{k=1}^p (s_{\omega_k} + 1) \log\left\{ 2^{2 L_{\omega_k} + 5}
    \delta^{-1} (L \omega_k+1) p_{\omega_k 0}^2 p_{\omega_k
      L_{\omega_k}+1}^2 s_{\omega_k}^{2L_{\omega_k}}\right\}\right].
\ese
Now plugging $\delta = n^{-1}$ using Condition
\ref{con:c2}--\ref{con:c4} and the fact that $d /n \to 0$ and $T/n\to
0$ and $p < \infty$, we have
\bse
&& \Pr\left\{\sup_{g \in \mF_{g}, \alpha_k\in \mH_k}|\Omega (g_0, \ba_0, \psi_{ g}, \bpsi_{\ba}, \epsilon_g^\dagger,
\beps_{\ba}^\dagger) - E\{\Omega (g_0, \ba_0, \psi_{ g}, \bpsi_{\ba}, \epsilon_g^\dagger,
\beps_{\ba}^\dagger) \} |  >Tp t + C_H T  p/n\right\}\\
&\leq& 2 \exp\left[- c_H n\min(t^2, t) + c_1 n\phi_{gn} L_g \log^2 (n)
+ \sum_{j=1}^p c_{2j} n\phi_{\omega_j n} L_{\omega_j} \log^2
(n)\right]. 
\ese

Let $t = \max\left[ 2 C_1 \log^2 (n) \left\{\phi_{gn} L_g  + \sum_{j=1}^p \phi_{\omega_j n} L_{\omega_j} \log^2
(n) \right\}, 1 \right]$, where $C_1 =  2 \{c_1 + \max_j(c_{2j})\}/c_H$, we have
\bse
&& \Pr\left\{\sup_{g \in \mF_{g}, \alpha_k\in \mH_k}|\Omega (g_0, \ba_0, \psi_{ g}, \bpsi_{\ba}, \epsilon_g^\dagger,
\beps_{\ba}^\dagger) - E\{\Omega (g_0, \ba_0, \psi_{ g}, \bpsi_{\ba}, \epsilon_g^\dagger,
\beps_{\ba}^\dagger) \} | \right.\\
&&\left. > 2Tp C_1 \log^2 (n) \left\{\phi_{gn} L_g  + \sum_{j=1}^p \phi_{\omega_j n} L_{\omega_j} \right\} + C_H T  p/n\right\} \\
&& \leq 2\exp \left[- n C_1 \log^2 (n) \left\{\phi_{gn} L_g  +
    \sum_{j=1}^p \phi_{\omega_j n} L_{\omega_j} \right\} \right]. 
\ese
Therefore, 
 \bse
 && |\Omega (g_0, \ba_0, \psi_{\wh g}, \bpsi_{\wh{\ba}}, \epsilon_g^\dagger,
\beps_{\ba}^\dagger) - E\{\Omega (g_0, \ba_0, \psi_{ \wh{g}}, \bpsi_{\wh{\ba}}, \epsilon_g^\dagger,
\beps_{\ba}^\dagger) \} | \\
&\leq&  Tp C_1 \log^2 (n) \left\{\phi_{gn} L_g  + \sum_{j=1}^p \phi_{\omega_j n} L_{\omega_j} \right\} + C_H T  p/n, 
 \ese
 with probability greater than $1 -  2\exp \left[- n C_1 \log^2 (n) \left\{\phi_{gn} L_g  +
    \sum_{j=1}^p \phi_{\omega_j n} L_{\omega_j} \right\} \right]$ for some
 $C_1>0$. Hence 
\bse
&& \Omega (g_0, \ba_0, \psi_{\wh g}, \bpsi_{\wh{\ba}}, \epsilon_g^\dagger,
\beps_{\ba}^\dagger) \geq  \alpha_{\min} E \left[\psi_{\wh{g}}\left\{\S_i(\ba_0)\right\}^2 + \sum_{j=1}^p\sum_{l=1}^d T
   \psi_{\wh{\alpha}_j}(\X_{il}, \X_i)^2\right]\\
&& -   Tp C_1 \log^2 (n) \left\{\phi_{gn} L_g  + \sum_{j=1}^p
   \phi_{\omega_j n} L_{\omega_j} \right\} -  C_H T  p/n
\ese
 with probability greater than $1 -  2\exp \left[- n C_1 \log^2 (n) \left\{\phi_{gn} L_g  +
    \sum_{j=1}^p \phi_{\omega_j n} L_{\omega_j} \right\} \right]$. 
This proves the result.

\section{Theorem \ref{th:supp1}}
\label{sec:supp:conv}

Below we derive the convergence rate and establish the asymptotic consistency of the estimator as a supplement to Theorem \ref{th:1} in the main material.
For the positive constants $C_0, A_g,
A_{\omega j}, a_g, a_{\omega j}, j = 1, \ldots, p$, we define
\bse
 \eta(A_g, a_g, \A_{\omega}, \a_{\omega}) 
&\equiv& 
\sqrt{C_0 \log(n)/(cn)} +
\left\{
A_g \max_{u = 0, \ldots, q_g} a_g^{-\frac{ \beta_{gu}^*}{t_{gu}}} n^{-\frac{ \beta_{gu}^*}{2\beta_{gu}^* + t_{gu}}}
\right. \\
& & \left.
+ 2C_M M_L\sqrt{d \min(s_g, Tp)}
\sup_{j=1, \ldots, p} A_{\omega_j}
\max_{u = 0, \ldots, q_{\omega_j}} a_{\omega_j}^{-\frac{ \beta_{\omega_j u}^*}{t_{\omega_j u}}} n^{-\frac{\beta_{\omega_j u}^*}{2\beta_{\omega_j u}^* + t_{\omega_j u}}}
\right\},
\ese
where $\A_{\omega} = (A_{\omega_j}, j = 1, \ldots, p)\trans$, and $\a_{\omega} = (a_{\omega_j}, j = 1, \ldots, p)\trans$.

\begin{Th}\label{th:supp1}
Let $\psi_{\wh g}= \wh{g} - g_0$ and $\psi_{\wh \alpha_j} = \wh{\alpha}_j - \alpha_{0j}$. 
Assume $\wh{\alpha}_j, \alpha_{0j} \in \mH_{j}$, $\omega^*_j \in \mG_{\omega_j}$,  $\wh{g}, g_0 \in \mF_g, $ and $g^* \in \mG_{g}$.
Assume $T, p<s_g$ and Conditions \ref{con:c1}--\ref{con:RE} hold.
Let $d{P_{\ba} (\cdot)}$ be the probability density function of $\S_i(\ba)$, and assume
$\|d{P_{\ba_0} (\z)}/d{P_{\ba^*} (\z)}\|_{\infty}\geq m_h>0$.
Furthermore, let $dP_{\X}(\cdot)$ be the probability density for $\X_{i}$.
Then there are positive constants $D_g,  d_g,  \D_{\omega} = (D_{\omega_j}),\d_{\omega} = (d_{\omega j}), D_{g 1}, d_{g1}, D_{\omega_j 1}, d_{\omega_j 1}, c_{H}', C_H$ such that 
\bse
&& \int \{\wh{g} (\z) - g^* (\z)\}^2 dP_{\ba^*}(\z) +  \sum_{j=1}^p \int
 \sum_{l=1}^d 
 \{\wh{\alpha}_{j}(\x_{l}, {\x})
 - \alpha^*_j (\x_{l}, {\x}) \} ^2 dP_{{\X}}({\x}) \\
%  &\leq& 
% \int \{\wh{g} (\z) - g^* (\z)\}^2 dP_{\ba^*}(\z) + T \sum_{j=1}^p \int
%  \sum_{l=1}^d 
%  \{\wh{\alpha}_{j}(\x_{l}, {\x})
%  - \alpha^*_j (\x_{l}, {\x}) \} ^2 dP_{{\X}}({\x}) \\
 &\leq& 16 \min(m_h, 1)^{-1}\alpha_{\min}^{-1}  \left(C_M M_L \eta(D_g, d_g,
   \D_{\omega}, \d_{\omega}) \left[ \min(s_g, Tp)^{1/2} \left\{
\log(n)/(cn)\right\}^{1/4} \right. \right. \\
&& \left. + \left\{\log(n)/(cn)\right\}^{1/4}  \right]
 \left.
+ Tp C_1 \log^2 (n) \left\{\phi_{gn} L_g  + \sum_{j=1}^p
  \phi_{\omega_j n} L_{\omega_j} \right\} +   C_H T  p/n
\right) \\
&& + 32 \min(m_h, 1)^{-1} C_M M_L dp \alpha_{\min}^{-2} \eta(D_g, d_g,
   \D_{\omega}, \d_{\omega})^2+ \left(D_{g1} \phi_{gn}+T\sum_{j=1}^p D_{\omega_j 1}
  \phi_{\omega_j n}\right)
\ese
with probability greater than $1-6n^{-1} -2\exp \left[- n C_1 \log^2 (n) \left\{\phi_{gn} L_g  + \sum_{j=1}^p \phi_{\omega_j n} L_{\omega_j} \right\} \right]$. 
When $n\to \infty$ and $T, p$ are finite, we have
\bse
&&\int \{\wh{g} (\z) - g^* (\z)\}^2 dP_{\ba^*}(\z) +  \sum_{j=1}^p \int
 \sum_{l=1}^d 
 \{\wh{\alpha}_{j}(\x_{l}, {\x})
 - \alpha^*_j (\x_{l}, {\x}) \} ^2 dP_{{\X}}({\x}) \overset{p}{\rightarrow}  0.
\ese

\end{Th}
The upper bound of the error is dominated by $Tp C_1 \log^2(n) \left\{\phi_{gn} L_g + \sum_{j=1}^p \phi_{\omega_j n} L_{\omega_j} \right\}$ when $T$ and $p$ are finite. Also because $\phi_{gn}$ and $\phi_{\omega_j n}$ are of the order of the inverse of polynomial functions of $n$, the upper bound approaches zero with the probability increasing to one as the sample size $n \to \infty$.
This shows that the error converges to zero in probability.

% The proof of the theorem is presented in Section \ref{sec:pro:th1} in the
% supplementary material. To prove the theorem, Lemma
% \ref{lem:firstorder} shows that the first derivative of the objective
% function has an upper bound that approaches zero as the sample size
% increases; Lemma \ref{lem:secondorder} then shows that the second
% derivative is asymptotically a positive definite matrix. We then
% derive the final results by combining these findings with the
% convergence of the deep learning function presented in Lemma
% \ref{lem:hh}.
%%%

\section{Proof of Theorem \ref{th:supp1}}\label{sec:pro:th1}
\noindent Proof: Because $\wh{g}$, $\wh{\ba}$ is the minimizer of
(\ref{eq:loss}), we have
\bse
0 &\geq & \mL(\wh{g}, \wh{\ba}) - \mL(g_0,  \ba_0)\\
&=& -n^{-1}
\sumi\left\{\mystrut\left( Y_i - h'\left[g_0\left \{\S_i(\ba_0) \right\}\right]\right) \right.\nonumber\\
&&\left. \sum_{j=1}^p \sum_{t=1}^T\nabla_{jt}  g_0\left \{\S_i(\ba_0) \right\} \sum_{l =1}^d 
  (X_{ilt} + Z_{ilj})\psi_{\wh{\alpha}_j}(\X_{il}, \X_i)\right\} \\
&& - n^{-1}
\sumi\left[\left( Y_i - h'\left[g_0\left \{\S_i(\ba_0) \right\}\right]
  \right) \psi_{\wh g} \left \{\S_i(\ba_0) \right\}\right]\\
&& + \Omega (g_0, \ba_0, \psi_{\wh g}, \bpsi_{\wh{\ba}}, \epsilon_g^\dagger,
\beps_{\ba}^\dagger)/2, 
\ese
where $\epsilon_g^\dagger, \epsilon_{\alpha_j}^\dagger \in (0, 1)$ and
$\beps_{\ba}^\dagger = (\epsilon_{\alpha_j}^\dagger, j = 1, \ldots,
p)\trans$.  By Lemma \ref{lem:firstorder} and \ref{lem:secondorder},
we have there are positive constants $D_g, d_{g}, \D_{\omega} = (D_{\omega_j}),
\d_{\omega} = (d_{\omega_j})$ such
that
\bse
&& 4 C_M M_L\eta(D_g, d_g,
   \D_{\omega}, \d_{\omega})\left[ \left[ dp E
\left\{\sum_{j=1}^p \sum_{l=1}^d T \psi_{\wh{\alpha}_j}(\X_{il},
  \X_i)^2\right\} \right]^{1/2} \right. \\
  && \left. + 
\left(E\left[ \psi_{\wh g} \left\{\S_i(\ba_0)\right\}^2\right]\right)^{1/2}\right]
+4 C_M M_L  \eta(D_g, d_g,
   \D_{\omega}, \d_{\omega}) \left[ \min(s_g, Tp)^{1/2}  \left\{
\log(n)/(cn)\right\}^{1/4}  \right. \\
&& \left. + \left\{\log(n)/(cn)\right\}^{1/4}  \right]
+  Tp C_1 \log^2 (n) \left\{\phi_{gn} L_g  + \sum_{j=1}^p
  \phi_{\omega_j n} L_{\omega_j} \right\}/2 +   C_H T  p/ (2n) \\
& \geq & \alpha_{\min} E \left[\psi_{\wh{g}}\left\{\S_i(\ba_0)\right\}^2 + \sum_{j=1}^p\sum_{l=1}^d
   T \psi_{\wh{\alpha}_j}(\X_{il}, \X_i)^2\right]/2 
\ese
with probability greater than $1-6n^{-1} -2\exp \left[- n C_1 \log^2 (n) \left\{\phi_{gn} L_g  +
    \sum_{j=1}^p \phi_{\omega_j n} L_{\omega_j} \right\} \right]$. This
implies
\bse
&&4C_MM_L \eta(D_g, d_g,
   \D_{\omega}, \d_{\omega})  (dp /2)^{1/2}\left(E
\left\{\sum_{j=1}^p\sum_{l=1}^d T \psi_{\wh{\alpha}_j}(\X_{il},
  \X_i)^2\right\} \right.\\
&&\left.+ E
\left[ \psi_{\wh g} \left\{\S_i(\ba_0) \right\}^2\right]\mystrut\right)^{1/2} \\
&&+ 4C_MM_L  \eta(D_g, d_g,
   \D_{\omega}, \d_{\omega}) \left[ \min(s_g, Tp)^{1/2} \left\{ 
\log(n)/(cn)\right\}^{1/4}  + \left\{16 \log(n)/(cn)\right\}^{1/4}  \right]\\
&&+  Tp C_1 \log^2 (n) \left\{\phi_{gn} L_g  + \sum_{j=1}^p
  \phi_{\omega_j n} L_{\omega_j} \right\}/2+   C_H T  p/ (2n) \\
&& \geq \alpha_{\min} E \left[\psi_{\wh{g}}\left\{\S_i(\ba_0)\right\}^2 + \sum_{j=1}^p\sum_{l=1}^d
   T \psi_{\wh{\alpha}_j}(\X_{il}, \X_i)^2\right]/2. 
\ese
Therefore, we have 
 \bse
&&  E \left[\psi_{\wh{g}}\left\{\S_i(\ba_0)\right\}^2 + \sum_{j=1}^p\sum_{l=1}^d
   T \psi_{\wh{\alpha}_j}(\X_{il}, \X_i)^2\right]\\
 &\leq & 
16 \alpha_{\min}^{-1}  \left(C_M M_L \eta(D_g, d_g,
   \D_{\omega}, \d_{\omega}) \left[ \min(s_g, Tp)^{1/2} \left\{
\log(n)/(cn)\right\}^{1/4}  + \left\{\log(n)/(cn)\right\}^{1/4}  \right] \right.\\
&&\left.+  Tp C_1 \log^2 (n) \left\{\phi_{gn} L_g  + \sum_{j=1}^p
  \phi_{\omega_j n} L_{\omega_j} \right\}/2 +   C_H T  p/ (2n)
\right)\\
&& + 32C_M M_L dp \alpha_{\min}^{-2} \eta(D_g, d_g,
   \D_{\omega}, \d_{\omega})^2.  
 \ese
 Now because $\|d{P_{\ba_0} (\z)}/d{P_{\ba^*} (\z)} \|_{\infty}\geq m_h$ almost surely , we obtain
 \bse
 &&  E \left[\psi_{\wh{g}}\left\{\S_i(\ba^*)\right\}^2 + \sum_{j=1}^p\sum_{l=1}^d
 T  \psi_{\wh{\alpha}_j}(\X_{il}, \X_i)^2\right] \\
 &\leq& 
16 \min(m_h, 1)^{-1}\alpha_{\min}^{-1}  \left(C_M M_L \eta(D_g, d_g,
   \D_{\omega}, \d_{\omega}) \left[ \min(s_g, Tp)^{1/2} \left\{
\log(n)/(cn)\right\}^{1/4} \right.\right. \\
&& \left.\left.
+ \left\{\log(n)/(cn)\right\}^{1/4}  \right] 
+  Tp C_1 \log^2 (n) \left\{\phi_{gn} L_g  + \sum_{j=1}^p
  \phi_{\omega_j n} L_{\omega_j} \right\} +   C_H T  p/n
\right)\\
&& + 32 \min(m_h, 1)^{-1} C_M M_L dp \alpha_{\min}^{-2} \eta(D_g, d_g,
   \D_{\omega}, \d_{\omega})^2.  
\ese
Now again using Lemma \ref{lem:fdistance},  we obtain there are constants
$D_{g 1}, 
D_{\omega_j 1}$ such that 
\bse
 && \int \{\wh{g} (\z) - g^* (\z)\}^2 dP_{\ba^*}(\z) + T \sum_{j=1}^p \int
 \sum_{l=1}^d 
 \{\wh{\alpha}_{j}(\x_l, {\x})
 - \alpha^*_j (\x_l, {\x}) \} ^2 dP_{\X}(\x) \\
 &\leq& 16  \min(m_h, 1)^{-1}\alpha_{\min}^{-1}  \left(C_M M_L \eta(D_g, d_g,
   \D_{\omega}, \d_{\omega}) \left[ \min(s_g, Tp)^{1/2} \left\{
\log(n)/(cn)\right\}^{1/4} \right.\right. \\
&& \left.\left.
+ \left\{\log(n)/(cn)\right\}^{1/4}  \right] 
+  Tp C_1 \log^2 (n) \left\{\phi_{gn} L_g  + \sum_{j=1}^p
  \phi_{\omega_j n} L_{\omega_j} \right\} +   C_H T  p/n
\right)\\
&& + 32 \min(m_h, 1)^{-1} C_M M_L dp \alpha_{\min}^{-2} \eta(D_g, d_g,
   \D_{\omega}, \d_{\omega})^2+ \left(D_{g1} \phi_{gn}+T\sum_{j=1}^p D_{\omega_j 1}
  \phi_{\omega_j n}\right)
\ese
with probability greater than $1-6n^{-1} -2\exp \left[- n C_1 \log^2 (n) \left\{\phi_{gn} L_g  +
    \sum_{j=1}^p \phi_{\omega_j n} L_{\omega_j} \right\}
\right]$. This proves the result.

\section{Examples of Theorem \ref{th:supp1}}
\label{sec:supp:ex}

We discuss the convergence rates under special cases of $g$, $\omega_k, k=1, \ldots, p$.
This discussion is similar to Section 4 in \cite{schmidt2020nonparametric}.

\textit{Additive model}:
Assume that $\omega_k, k = 1, \ldots, p$ has the form
$$
\omega_k(\x) = \sum_{t=1}^T \omega_{kt} (x_t).
$$
This can be written as a composition of functions
$$
\omega_{k} = \omega_{1k} \circ \omega_{0k}
$$
where $\omega_{0k}(\x) = (\omega_{0k1}(x_1), \ldots, \omega_{0kT}(x_T))\trans$
and $\omega_{1k}({\bf y}) = \sum_{t=1}^T y_t$.
Then $\omega_{0k}: [-1, 1]^T \to \bR^T$ and $\omega_{1k}: \bR^T \to \bR$.
Thus, for $\omega_k \in \mG_{\omega_k}$ we have $q_{\omega_k}=1$, ${\bf d}_{\omega_k} = (d_0, d_1, d_2) = (T, T, 1)$, and ${\bf t}_{\omega_k} = (t_0, t_1) = (1, T)$.
Suppose $\omega_{kt} \in \mC_1^\beta ([-1, 1], K)$ for $t = 1, \ldots, T$.
Then $\omega_k: [-1, 1]^T \overset{\omega_{0k}}{\longrightarrow} [-K, K]^T \overset{\omega_{1k}}{\longrightarrow} [-KT, KT]$.
Since for any $\beta_1 > 1$, $\omega_{1k} \in \mC_T^{\beta_1} ([-K, K]^T, (K+1)T)$, then $\phi_{\omega_jn} = n^{-\frac{2\beta}{2\beta+1}}$ and
$$
\omega_k \in \mG(1, (T, T, 1), (1, T), (\beta, \max(\beta, 2) T), (K+1) T).
$$

Similarly, assume that $g$ has the form
$$
g(\z) = \sum_{k=1}^p \sum_{j=1}^T g_{jk} (z_{jk}).
$$
This can be written as a composition of functions
$$
g = g_1 \circ g_0
$$
where $g_0(\z) = (g_{jk}(z_{jk}), j=1, \ldots, T; k=1, \ldots, p)_{T\times p}$ and $g_1({\bf y}) =  \sum_{k=1}^p \sum_{j=1}^T y_{jk}$.
Thus for $g \in \mG_g$ we have $q_g = 1$, $\d_g = (d_0, d_1, d_2) = (Tp, Tp, 1)$, and $\t_g = (t_0, t_1) = (1, Tp)$.
Suppose $g_{jk} \in \mC_1^\beta ([-d, d], K)$ for $j = 1, \ldots, T$ and $k = 1, \ldots, p$.
Then $g: [-d, d]^{T\times p} \overset{g_0}{\longrightarrow} [-K, K]^{T\times p} \overset{g_1}{\longrightarrow} [-KTp, KTp]$.
Thus, $\phi_{gn} = n^{-\frac{2\beta}{2\beta+1}}$ and
$$
g \in \mG(1, (Tp, Tp, 1), (1, Tp), (\beta, \max(\beta, 2)Tp), (K+1)Tp).
$$

Then for network architectures $\mF_{\omega_j}, \mF_g$ with $L_{\omega_jn} \lesssim \log n$, $L_g \lesssim \log n$ satisfying Conditions \ref{con:c1}--\ref{con:c4},
we have
$F_{\omega_j} \ge (K+1)T$, $F_g \ge (K+1)Tp$, $2\log_2 (4\max (\beta, 2) T) \log n \le L_{\omega_j} \lesssim \log n$, $2\log_2 (4 \max(\beta, 2) Tp) \log n \le L_g \lesssim \log n$,
$n^{1/(2\beta+1)} \lesssim \min_u p_{\omega_ju}, j = 1, \ldots, p$, $n^{1/(2\beta+1)} \lesssim \min_u p_{gu}$,
$s_{\omega_j} \asymp n^{1/(2\beta+1)} \log n$, $s_g \asymp n^{1/(2\beta+1)} \log n$.
Through Theorem \ref{th:supp1}, under Conditions \ref{con:c1}--\ref{con:RE}, we obtain
$$
Tp C_1 \log^2(n) \left\{\phi_{gn} L_g + \sum_{j=1}^p \phi_{\omega_j n} L_{\omega_j} \right\}
\lesssim
Tp(p+1)C_1 n^{-\frac{2\beta}{2\beta+1}} \log^3 n.
$$
% Then the estimation error is upper bounded by the $\log^3n$-factor with probability greater than $1 - 6n^{-1} - 2\exp \left\{-C_1 \left[ 2(p+1) \log_2 (4\max(\beta, 2)T) + 2 \log_2 p \right] n^{\frac{1}{(2\beta+1)}} \log^3n \right\}$.
This shows that the estimators converge at least at the rate $\log^3 n$.

%%%%%%%%%%%%%%%%%%%%%%%%%%%%%%%%%%%%%%%%%%%%%%%%%%%%%%%%%%%%%%%%%%%%%%%%%%%%%%

\section{Simluation}
\label{sec:simu}

We evaluate NDL using simulation studies with different sample sizes $n$ to examine the convergence of the estimation procedure.

To simulate data that closely resemble real-world conditions, we randomly segment $d \times (T + p)$ brain signal matrices from the TUH data, with  $d = 22$, $T = 64$ and $p=64$. Each matrix is standardized by subtracting the mean of its rows and dividing by the standard deviation of its elements. 
We then extract the segment between the $(p/2 + 1)$th column and the $(p/2 + T)$th column as $\X_i$, forming a $d \times T$ matrix. The remaining columns are used as $\Z_i$, a $d \times p$ matrix. We choose the true functions $\omega_k^*(\X_{il}),\; k = 1, \ldots, p,$ randomly with replacement from the following set:
\bse
&
\left\{
\log \left( \sum_{t=1}^T (X_{ilt})^2 / T -\mu_{il}^2 \right),
-\log \left( \sum_{t=1}^T (X_{ilt})^2 / T -\mu_{il}^2 \right), \right.\\
&
\left( \sum_{t=1}^T (z_{ilt})^3 / T \right),
-\left( \sum_{t=1}^T (z_{ilt})^3 / T \right),
\left( \sum_{t=1}^T (z_{ilt})^4 / T - 3 \right),
-\left( \sum_{t=1}^T (z_{ilt})^4 / T - 3 \right), \\
&\left.
\log \left( \sum_{t=1}^T |X_{ilt} \sin(X_{ilt})| \right),
\log  \left( \sum_{t=1}^T |X_{ilt} \cos(X_{ilt})| \right)
\right\},
\ese
where $z_{ilt} = (X_{ilt} - \mu_{il}) / \sigma_{il}$ with $\mu_{il} = \sum_{t=1}^T X_{ilt} / T$ and $\sigma_{il} = \sqrt{\sum_{t=1}^T (X_{ilt})^2 / T -\mu_{il}^2}$.
Furthermore, we generate $\bb_1 = (\beta_{11}, \ldots, \beta_{1T})\trans$ and $\bb_2 = (\beta_{21}, \ldots, \beta_{2p})\trans$ from the standard multivariate normal distributions and $\beta_0$ from the standard normal distribution. The coefficient parameters $\{\beta_0,\bb_1, \bb_2\}$ are fixed for all simulation repetitions.
Finally, we define
$ g^*\left\{\S_i(\ba^*)\right\}= \bb_1\trans \S_i(\ba^*) \bb_2 + \beta_0 $, and sample $Y_i$ as a Bernoulli variable with the probability of success being $(\exp[-g^*\left\{\S_i(\ba^*)\right\}] + 1)^{-1}$.

We apply NDL to each simulated sample set $\{ (\X_i, \Z_i, Y_i), i = 1, \ldots, n \}$ and obtain the resulting estimators $\wh\ba, \wh g$.
Then we measure the estimation errors of $\wh\ba$ and $\wh g$, respectively:
\bse
d_{\ba^*}(\wh\ba) = \frac{1}{N} \sum_{i=1}^N
\| \{\wh\ba(\X_{il}, \X_i), l=1,\ldots,d\} \trans
- \{\ba^*(\X_{il}, \X_i), l=1,\ldots,d\} \trans \|_2,
\ese 
and 
\bse
d_{g^*}(\wh g) = \frac{1}{N} \sum_{i=1}^N
| h'[\wh g\{\S_i(\wh\ba)\}] - h'[g^* \{\S_i(\wh\ba)\}] |, 
\ese
where  $N = 4096$ samples are randomly selected from simulated data to evaluate the estimation errors of the functions.

\begin{figure}[htbp]

\begin{subfigure}{.5\textwidth}
\centering
\includegraphics[page=1,width=1.0\linewidth]{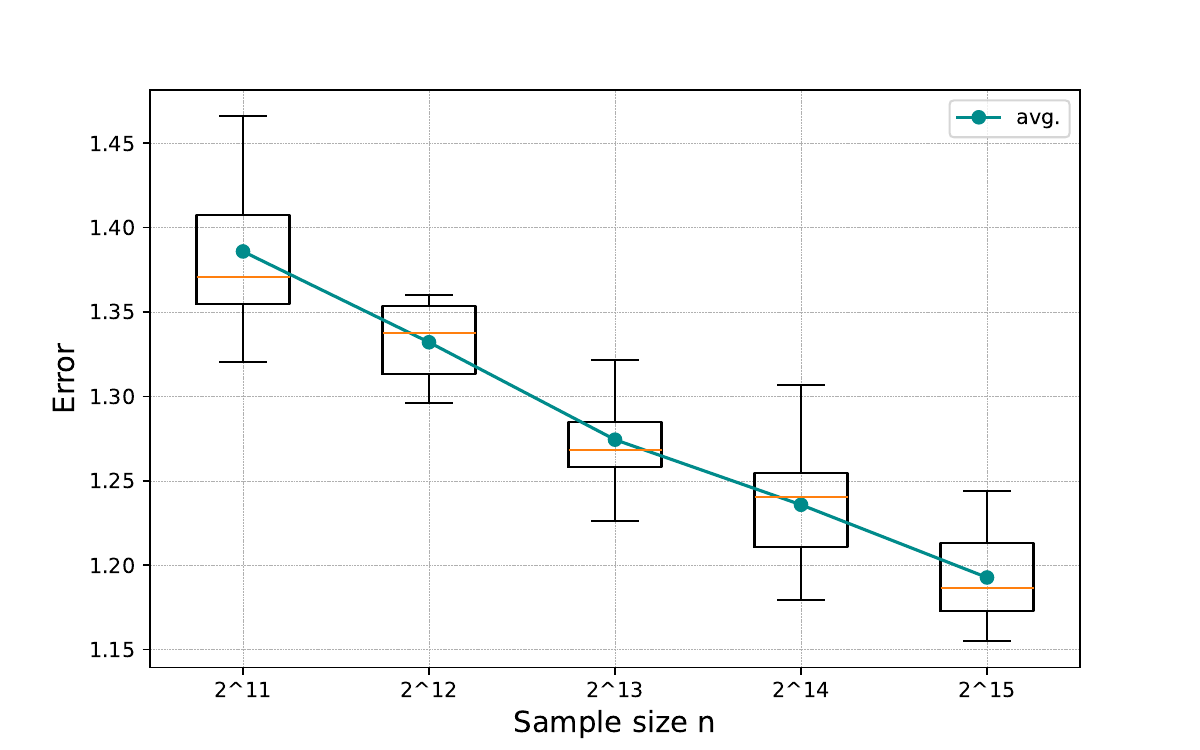}
\caption{$d_{\ba^*}(\wh\ba)$}
\vspace{-2mm}
\end{subfigure}
\begin{subfigure}{.5\textwidth}
\centering
\includegraphics[page=2,width=1.0\linewidth]{fig/simulation.pdf}
\caption{$d_{g^*}(\wh g)$}
\vspace{-2mm}
\end{subfigure}
\caption{Boxplots of the estimation errors under different sample sizes over 10 repetitions.
\label{fig:sim}}

\end{figure}

We generate data for five settings of varying sample sizes $n_j = 2^{10 + j}$, $j=1, \ldots, 5$. 
Each setting is repeated 10 times.
Figure~\ref{fig:sim} shows that $ d_{\ba^*}(\wh\ba)$
and $ d_{g^*}(\wh g)$ decrease as the sample size $n$ increases.
One can see that both $\ba(\X_{il}, \X_i)$ and $g\left\{\S_i({\ba})\right\}$ are identifiable and our estimators converge to the true values as the sample size increases.

\section{The Standard ACNS Temporal Central Parasagittal (TCP) Montage}

\begin{table}[htbp]
\setlength{\tabcolsep}{15pt} % Default value: 6pt
\renewcommand{\arraystretch}{0.8} % Default value: 1
\caption{
The standard ACNS TCP montage. \label{tab:TCP}}
\vspace{-5mm}
\begin{center}
\begin{tabular}[width=1.0\textwidth]{c|c|cc}
\hline
Index & Montage & Anode & Cathode \\
\hline
0 & FP1-F7 & EEG FP1-REF & EEG F7-REF \\
1 & F7-T3 & EEG F7-REF & EEG T3-REF \\
2 & T3-T5 & EEG T3-REF & EEG T5-REF \\
3 & T5-O1 & EEG T5-REF & EEG O1-REF \\
4 & FP2-F8 & EEG FP2-REF & EEG F8-REF \\
5 & F8-T4 & EEG F8-REF & EEG T4-REF \\
6 & T4-T6 & EEG T4-REF & EEG T6-REF \\
7 & T6-O2 & EEG T6-REF & EEG O2-REF \\
8 & A1-T3 & EEG A1-REF & EEG T3-REF \\
9 & T3-C3 & EEG T3-REF & EEG C3-REF \\
10 & C3-CZ & EEG C3-REF & EEG CZ-REF \\
11 & CZ-C4 & EEG CZ-REF & EEG C4-REF \\
12 & C4-T4 & EEG C4-REF & EEG T4-REF \\
13 & T4-A2 & EEG T4-REF & EEG A2-REF \\
14 & FP1-F3 & EEG FP1-REF & EEG F3-REF \\
15 & F3-C3 & EEG F3-REF & EEG C3-REF \\
16 & C3-P3 & EEG C3-REF & EEG P3-REF \\
17 & P3-O1 & EEG P3-REF & EEG O1-REF \\
18 & FP2-F4 & EEG FP2-REF & EEG F4-REF \\
19 & F4-C4 & EEG F4-REF & EEG C4-REF \\
20 & C4-P4 & EEG C4-REF & EEG P4-REF \\
21 & P4-O2 & EEG P4-REF & EEG O2-REF \\
\hline
\end{tabular}
\end{center}
\vspace{-5mm}
\end{table}

\newpage
\vspace{10mm}

\bibliographystyle{agsm-asa}

\bibliography{Bibliography-MM-MC}

\end{document}